\documentclass{fairmeta}
\usepackage{amsthm,amsmath,amssymb}
\usepackage{graphicx}
\usepackage{natbib}
\usepackage{subcaption}
\usepackage{algorithm,algorithmicx,algpseudocode}
\usepackage{pifont}

\usepackage{dblfloatfix}
\usepackage{enumitem}
\usepackage{booktabs}
\usepackage{multirow}
\usepackage{rotating}
\usepackage{bbding}
\usepackage{adjustbox}
\usepackage{xspace}
\usepackage{listings}
\usepackage{xcolor}
\usepackage{colortbl}
\usepackage{wrapfig}

\newcommand{\best}[1]{\textbf{#1}}

\title{RefVideo-6M: A Reliable Reference-Based Dataset for Instructional Video Editing}
\author[1,2,*]{Bojia Zi}
\author[1,*]{Xiaoyan Yang}
\author[1,3]{Yu Zhou}
\author[1,4]{Ruijie Sun}
\author[1,5]{Lihan Zhang}
\author[2]{Bin Liang}
\author[2,\dagger]{Kam-Fai Wong}
\author[1]{Haibin Huang}
\author[1]{Chi Zhang}
\author[1,\dagger]{Xuelong Li}
\affiliation[1]{Institute of Artificial Intelligence, China Telecom (TeleAI)}
\affiliation[2]{The Chinese University of Hong Kong (CUHK)}
\affiliation[3]{Sun Yat-sen University}
\affiliation[4]{Fudan University}
\affiliation[5]{Tsinghua University}
\contribution[*]{Equal Contribution}
\contribution[\dagger]{Corresponding Author}

\begin{document}

\abstract{
Recent advances in video editing have been largely driven by large-scale instruction-based datasets. However, existing datasets still suffer from two critical limitations. First, target videos are commonly produced by automatic editing models, which may introduce visible artifacts and unreliable supervision signals. Second, most public datasets rely primarily on textual instructions, while lacking visual references that are crucial for precise, identity-preserving, and controllable editing. To address these limitations, we introduce RefVideo-6M, a large-scale reference-guided editing dataset containing 5 million video editing samples and 1 million image editing samples. 
To ensure reliable supervision, our dataset uses a construction pipeline that treats artifact-free real videos as editing targets and generates quality-filtered input conditions with multiple editing experts. In addition, it provides approximately 6 million visual references, covering diverse reference types and editing scenarios, thereby enabling models to learn fine-grained visual correspondence beyond text-only instructions. Based on RefVideo-6M, we further train a reference-guided video editing model, Ref-MoT, to evaluate the effectiveness and scalability of the proposed dataset. Extensive experiments demonstrate that RefVideo-6M provides substantially more reliable supervision than existing datasets and enables the training of powerful editing models with improved visual quality, controllability, and reference consistency. \emph{The open-source dataset is available at \url{https://huggingface.co/datasets/RefVideo6M/RefVideo6M}.}
}

\maketitle

\begin{figure*}[!t]
\vspace{-1em}
    \centering
    \includegraphics[width=0.965\textwidth]{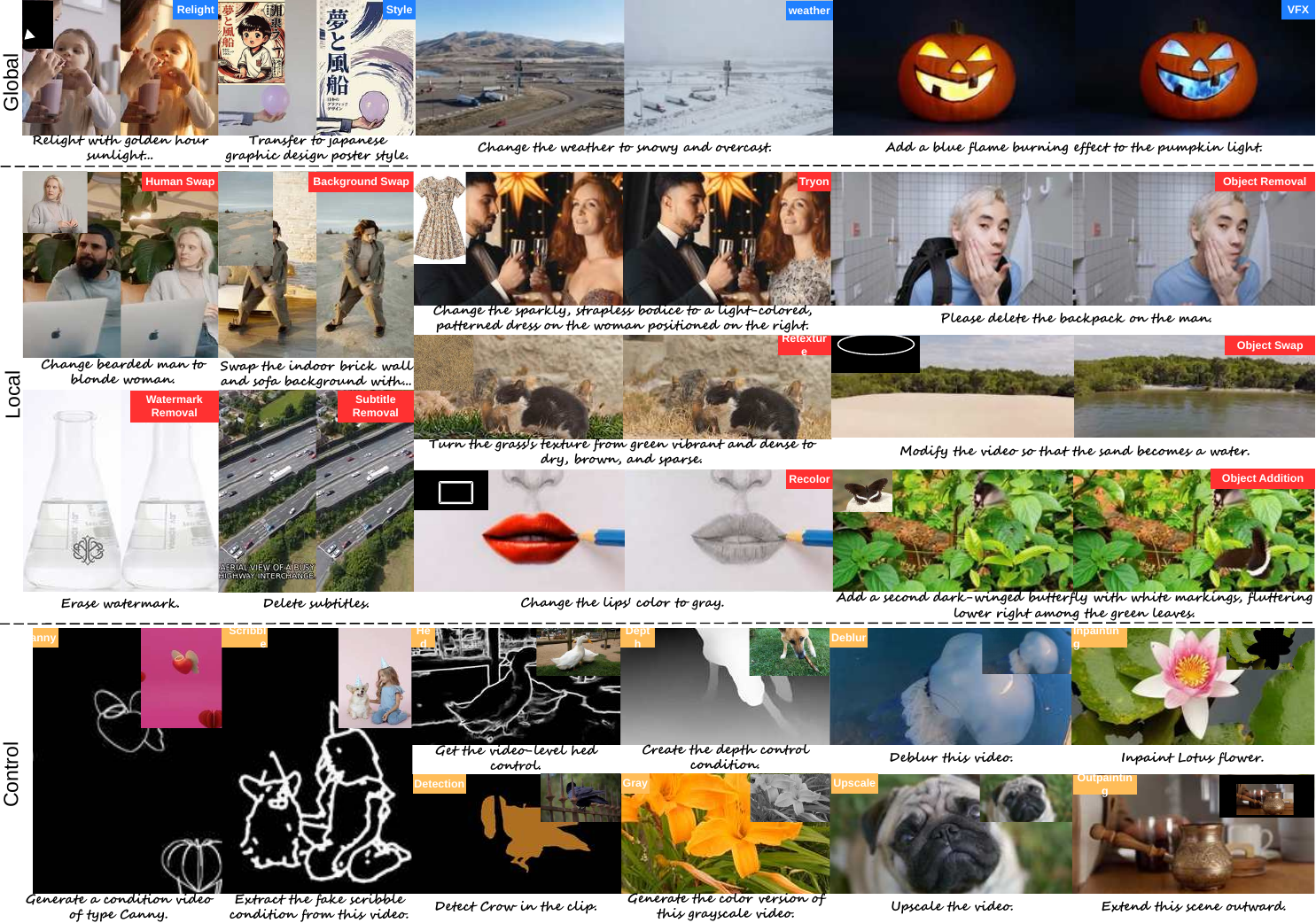}
    \vspace{-0.65em}
    \caption{Visualization of RefVideo-6M. Our dataset is a reliable reference-based video editing dataset. It provides a diverse set of reference types, including circles, bounding boxes, light directions, outpainting masks, background images, styles, textures, objects, humans, and clothing references. More visualization results can be found in the supplementary file.}
    \label{fig:teasers}
    \vspace{-1.5em}
\end{figure*}

\section{Introduction}

Video editing has achieved remarkable 
progress~\cite{dreamomni3_xia2025dreamomni3scribblebasededitinggeneration, omniedit_wei2024omniedit, flux2_klein, brushnet_ju2024brushnetplugandplayimageinpainting, powerpaint_zhuang2024task, hq_edit_hui2024hq, ultraedit_zhao2024ultraedit, step1x_edit_liu2025step1x_edit, qwenedit_wu2025qwenimagetechnicalreport, opens2v_yuan2025opens2v, chen2024videocrafter2, klingo1_net, wang2024videocomposer, hunyuanvideo_1_5_wu2025hunyuanvideo15technicalreport, cogvideox_yang2024cogvideox, gen4_aleph, veo3, an2026ai, shao2025ai,ku2024anyv2v, liangsen_2025omniv2v, insv2v_cheng2024consistent, wu2025insvie, chen2025hunyuanvideo, senorita_zi2025se, ditto_bai2025scalinginstructionbasedvideoediting, unic_ye2025unicunifiedincontextvideo, liangsen2026cot,
liangsen2024hypercorrelation, viva_cong2025vivavlmguidedinstructionbasedvideo, vino_chen2026vino, opensora, yuzhou_zhou2026point2insertvideoobjectinsertion, minimax_h3, 
liangsen2026spongebob}. 
Previously, video editing methods were primarily based on DDIM inversion~\cite{ddim_based_gal2022image}, 
which reconstructs the latent representation of a given video and modifies the prompt to regenerate edited results. 
More recent approaches have dropped inversion techniques and instead train dedicated editing models on large-scale editing datasets. 
InsV2V~\cite{insv2v_cheng2024consistent} represents an early instruction-based editing model built using datasets generated 
by VideoP2P~\cite{liu2024video}. 
EditVerse~\cite{editverse_ju2025editverseunifyingimagevideo} proposes a unified framework capable of both video generation 
and editing. 
VideoCoF~\cite{videocof_yang2025unifiedvideoeditingtemporal} adopts a Chain-of-Thought–inspired strategy to reason about more coherent editing results across frames. ICVE~\cite{icve_liao2025incontextlearningunpairedclips} constructs in-context video editing models using unpaired videos and fine-grained datasets, leveraging the full-DiT architecture~\cite{fulldit_ju2025fullditmultitaskvideogenerative}. Despite their algorithmic innovations, these methods also devote effort to editing data preprocessing and construction, highlighting the critical role of editing datasets in video editing models.

Recently, several video editing datasets have been introduced. 
Senorita-2M consists of 18 editing tasks with approximately 2M video editing pairs. InsViE contains 1M video editing pairs generated using Stable Video Diffusion ~\cite{stablevideodiffusion_blattmann2023stable} and further filtered by optical flow and GPT-4o~\cite{openai2023gpt4}. Ditto~\cite{ditto_bai2025scalinginstructionbasedvideoediting} constructs higher-resolution and longer-duration video editing data by leveraging VACE~\cite{vace_jiang2025vace} together with edited first frames. OpenVE-3M~\cite{openve3m_he2025openve3mlargescalehighqualitydataset} and ReCo~\cite{reco_zhang2025regionconstraintincontextgenerationinstructional} adopt similar data construction pipelines and demonstrate state-of-the-art performance on their respective editing benchmarks. Recently, Goku-2M~\cite{liang2026goku}, a concurrent work, proposed 10 editing tasks and constructed a dataset comprising 2 million video editing samples. \textbf{\emph{However, existing video editing datasets remain unreliable.}}
Most datasets rely on open-source editing models to generate edited videos, which often introduce artifacts even after LLM-based filtering. These artifacts weaken the effectiveness of training and ultimately limit the quality of the resulting editing models. Common issues such as low resolution, blurriness, and unintended side effects further impede the development of robust video editing methods.
\textbf{\emph{Moreover, most existing datasets support only textual prompts and lack visual references.}} This limitation restricts user interaction and reduces the controllability of editing models.

To address these issues, we propose RefVideo-6M, a large-scale dataset containing 5 million video editing pairs at 720p resolution with 81 to 129 frames, and 1 million image editing pairs. The video subset is constructed by 12 editing experts, while the image subset is generated with FLUX2-Klein-9B~\cite{flux2_klein}. Our dataset has two key advantages. \textit{First, our dataset is highly reliable.} For most editing tasks, we reverse the source and edited videos, using the edited video as the source and the original video as the target. \textit{Second, RefVideo-6M enables precise reference-based editing with visual prompts.} These references fall into two categories: region guidance, such as bounding boxes and circles, and visual references, such as style, texture, background, object, lighting direction, outpainting masks, and clothing images. 

Training on our dataset, a standard instruction-based editing model built on HunyuanVideo1.5, without any additional architectural innovations, achieves state-of-the-art performance among all baselines and variants. This demonstrates that our dataset effectively supports the development of high-quality editing models. To incorporate reference information into the editing model, we propose RefMoT, which delivers better editing results while reducing training computation by 50\% and maintaining an inference cost comparable to that of the standard editing model.

Our main contributions are summarized as follows:
\begin{enumerate}
    \item We introduce  \textbf{\textit{RefVideo-6M}}, a reliable large-scale video editing dataset with 5 million video pairs and 1 million image pairs, produced by 12 editing experts. We reverse the original and edited videos to build reliable video pairs and leverage LLM to filter out the failure cases. 
    
    \item RefVideo-6M consists of 6 million references to support reference-based editing. It covers 10 reference types, allowing users to point out both editing location and editing appearance.
  
    \item We train a simple instruction-based editing model and introduce the \textbf{\textit{RefMoT}} architecture to adapt it into a reference-based editing model. Comprehensive experiments demonstrate that our dataset and model architecture achieve state-of-the-art performance.
    
\end{enumerate}

\begin{figure*}[!ht]
    \vspace{-1em}
    \centering
    \includegraphics[width=1.0\linewidth]{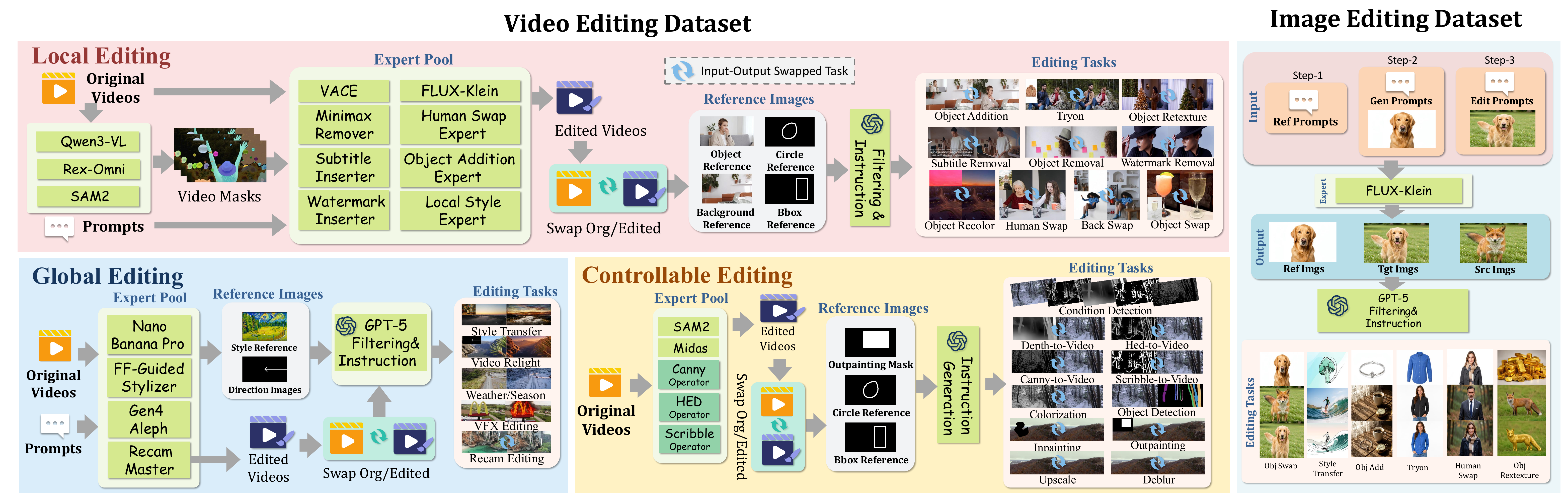}
    \vspace{-1.75em}
    \caption{Construction pipeline of RefVideo-6M. We train four experts in addition to eight open-source experts. In our dataset, most tasks reverse the roles of the original and edited videos to construct reliable video pairs.}
    \label{fig:data_pipeline}
\end{figure*}

\renewcommand{\arraystretch}{1.0}
\setlength{\tabcolsep}{4.85pt} 
\begin{table*}[!htp]\scriptsize
\centering 
\resizebox{\textwidth}{!}{%
\begin{tabular}{lccccccccc} 
\toprule \textbf{Datasets}  & \textbf{Experts} & \textbf{Tasks} & \textbf{Video Length} & \textbf{Resolution}  & \textbf{Ref-Img for Video}&\textbf{Video Pairs} &\textbf{Ref-Img for Image} &\textbf{Image Pairs} \\ \hline
InsV2V~\cite{insv2v_cheng2024consistent} & \ding{55}  & Free-Form & 16 & 256$^{2}$ & \ding{55}& 0.06M& \ding{55}& \ding{55}\\
InsViE~\cite{wu2025insvie}  & \ding{55} & Free-Form & 25 & 576P & \ding{55}& 1M& \ding{55}& \ding{55} \\   
Señorita~\cite{senorita_zi2025se}  & 8 & 16 Tasks & 33-64 & 336P/1120P & \ding{55}& 2M& \ding{55}& \ding{55}\\ 
Ditto~\cite{ditto_bai2025scalinginstructionbasedvideoediting}  & 3 & 11 Tasks & 101 & 720P & \ding{55}& 2M& \ding{55}& \ding{55}\\ 
FFP~\cite{ffp300k_huang2026ffp300kscalingfirstframepropagation}  & 2 & 5 Tasks& 81 & 720P & \ding{55}& 0.3M& \ding{55}& \ding{55}\\
Reco~\cite{reco_zhang2025regionconstraintincontextgenerationinstructional}  & 1 & 4 Tasks & 33-64 & 720P & \ding{55}& 0.5M& \ding{55}& \ding{55}\\ 
OpenVE~\cite{openve3m_he2025openve3mlargescalehighqualitydataset}  & 5 & 10 Tasks & 65-129 & 720P & \ding{55}& 2M& \ding{55}& \ding{55}\\ \hline
\textbf{RefVideo-6M} & \textbf{12} &\textbf{26 Tasks} & \textbf{81-129} & \textbf{720P} & \textbf{5M}& \textbf{5M}& \textbf{1M} & \textbf{1M} \\ \hline
\bottomrule
\end{tabular}
}
\caption{ Comparison among OpenVE-3M, Reco-500K, Ditto-1M, Señorita-2M, InsV2V and InsViE-1M. For the number of experts, we count only those based on deep learning and exclude traditional operators. For OpenVE-3M, only 2M samples are available in the released open-source dataset. For all datasets, the numbers of experts and tasks are recalculated according to our unified criteria.}
\vspace{-2em}
\label{tab:dataset_parameter_comparison} 
\end{table*}

\section{Related Works}
\subsection{Video Editing Datasets}

To produce instruction-based video editing models, reliable training data is essential. 
InsV2V first constructed a video editing dataset using synthetic video pairs generated by VideoP2P~\cite{liu2024video}, achieving state-of-the-art performance at that time. Subsequently, VIVID-10M~\cite{vivid_10m_hu2024vivid} was proposed for region-level editing; it employs inpainting experts trained on CogVideoX~\cite{cogvideox_yang2024cogvideox} to create a hybrid editing dataset consisting of both images and videos. Senorita-2M~\cite{senorita_zi2025se} utilizes a set of video editing experts to construct a dataset covering 18 video editing tasks, although the video pairs are at relatively low resolution. InsViE~\cite{wu2025insvie} leverages Stable Video Diffusion ~\cite{stablevideodiffusion_blattmann2023stable} to propagate the first source frame and edited frame, and further filters the generated results using optical flow and GPT-4o, resulting in a dataset of 1M video editing pairs; however, due to the limitations of Stable Video Diffusion, the resulting source and edited videos tend to be relatively static. Similarly, Ditto~\cite{ditto_bai2025scalinginstructionbasedvideoediting}, OpenVE-3M~\cite{openve3m_he2025openve3mlargescalehighqualitydataset}, and ReCo~\cite{reco_zhang2025regionconstraintincontextgenerationinstructional} construct video editing datasets using VACE~\cite{vace_jiang2025vace} and other editing models with guidance from edited first frames. Specifically, OpenVE-3M~\cite{openve3m_he2025openve3mlargescalehighqualitydataset} contains eight of the most common video editing tasks. Ditto~\cite{ditto_bai2025scalinginstructionbasedvideoediting} consists of 1M video pairs spanning two primary editing categories, including global editing and local editing, each further divided into a broad range of sub-tasks. ReCo~\cite{reco_zhang2025regionconstraintincontextgenerationinstructional} comprises 500K video editing pairs constructed using VACE. Similarly, FFP-300K~\cite{ffp300k_huang2026ffp300kscalingfirstframepropagation} is a dataset consisting of 300K video editing pairs generated through first-frame propagation strategies.

\textbf{However, these datasets rely on existing editing models to generate edited videos and treat them as ground truth targets, which may introduce artifacts into the annotations. Moreover, they typically accept only textual instructions as input, rather than supporting both textual and visual guidance.}

\subsection{Video Editing Methods}
In recent years, numerous video editing methods have been proposed. 
Unlike early inversion-based approaches~\cite{liu2024video, tuneavideo_wu2023tune, tokenflow_geyer2023tokenflow, cong2023flatten, ku2024anyv2v}, 
instruction-based editing methods require training but offer faster inference speed, simpler prompts, and improved visual quality~\cite{insv2v_cheng2024consistent, wu2025insvie, senorita_zi2025se, univideo_wei2026univideounifiedunderstandinggeneration, unic_ye2025unicunifiedincontextvideo, openve3m_he2025openve3mlargescalehighqualitydataset, lucy_edit_decart2025lucyedit, easyv2v_mai2025easyv2vhighqualityinstructionbasedvideo, odiscoedit_chen2025odiscoeditobjectdistortioncontrol, instructx_mou2025instructxunifiedvisualediting, viva_cong2025vivavlmguidedinstructionbasedvideo, revise_liu2025revisereasoninformedvideoediting, propfly_seo2026propflylearningpropagateonthefly,liang2026cot,feng2026mseditor,xie2026grnedit}. InsV2V~\cite{insv2v_cheng2024consistent} introduces Long Video Sampling Correction to maintain temporal consistency during long-sequence editing, while PropGen~\cite{propgen_liu2024generative} supervises frame-wise changes using masks of edited regions, reducing the need for large-scale datasets. FFP~\cite{ffp300k_huang2026ffp300kscalingfirstframepropagation} edits only the first frame and propagates the modifications to subsequent frames, and further proposes Adaptive Spatio-Temporal RoPE to better model spatio-temporal relationships. Ditto~\cite{ditto_bai2025scalinginstructionbasedvideoediting} builds an instruction-based editing model upon the VACE-14B~\cite{vace_jiang2025vace}. VideoCoF~\cite{videocof_yang2025unifiedvideoeditingtemporal} adopts a Chain-of-Frames paradigm inspired by Chain-of-Thought reasoning and trained on curated datasets to achieve high-quality editing results. Insvie~\cite{wu2025insvie} employs a multi-stage learning strategy to progressively improve instruction-following and editing capabilities. OpenVE~\cite{openve3m_he2025openve3mlargescalehighqualitydataset} introduces a Mixture-of-Experts connector to bridge multimodal inputs and visual hidden states for precise control. UNIC~\cite{unic_ye2025unicunifiedincontextvideo} performs in-context video editing using reference editing resources, while ICVE~\cite{icve_liao2025incontextlearningunpairedclips} leverages unpaired clips and fine-grained datasets to enhance editing performance and NOVA~\cite{nova_pan2026novasparsecontroldense} also get escape from the paired videos usage. 

Recently, unified models capable of video generation, editing, and understanding within a single framework have attracted significant attention. Capybara~\cite{capybara_2026raocapybara} is a unified visual creation model with video editing capabilities. UniVideo~\cite{univideo_wei2026univideounifiedunderstandinggeneration} integrates understanding, generation, and editing by conditioning a DiT backbone on multiple modalities, built upon HunyuanVideo~\cite{hunyuanvideo_kong2024hunyuanvideo}. Similarly, Omni-Video-2~\cite{omnivideo2_yang2026omnivideo2scalingmllmconditioned} performs unified generation and editing with MLLM-based conditioning. More recent unified frameworks~\cite{viva_cong2025vivavlmguidedinstructionbasedvideo, vino_chen2026vino, mamMoTh2_5} further improve both generation and editing quality. EditVerse~\cite{editverse_ju2025editverseunifyingimagevideo} combines curated generation and editing datasets to train a single model capable of handling diverse tasks, leveraging self-attention for robust in-context learning, cross-modal knowledge transfer, and flexible processing of inputs and outputs with arbitrary resolutions and durations. 

\textbf{Overall, the performance of video editing models is highly dependent on the quality of training data, and for a given model architecture, higher-quality editing datasets generally lead to stronger editing performance.}

\section{Methodology}

\subsection{Dataset Construction}

Our dataset consists of 5 million video samples covering 26 video editing tasks,
 and 1 million image samples covering 6 image editing tasks. 
 The construction pipeline is shown in Figure \ref{fig:data_pipeline}.

\noindent \subsubsection{Dataset Preparation}

We collected approximately 1M videos from the Pexels.com with their permissions. We first removed the duplicate samples by comparing the similarities between features from different videos encoded by DINO-V3 ~\cite{dinov3}. 
We further filtered out videos shorter than 161 frames. After filtering, we retained around 700K videos, which were split into 2.5M segments, each containing 161 frames. We then used Qwen-VL-3-8B~\cite{Qwen3-VL} to extract captions and object names. The extracted object names were provided to Rex-Omni~\cite{rex_omni_jiang2025detectpointprediction} to obtain bounding boxes in the first frame. Finally, we fed the bounding boxes and videos into SAM-2~\cite{sam2_ravi2025sam2} to generate object masks. In total, we obtained approximately 40M masks over 2.5M video segments, each paired with object names and captions.

\subsubsection{Reliable Video Editing Dataset}
Our video dataset consists of three categories, including global editing, local editing and controllable editing.

\noindent \textit{\textbf{Global Editing}}. In this category, we have global stylization, relight, recam, weather/season editing and visual effect editing. We train a global editor, which can propagate the changes in the first frame to the rest frames. This editor is based on Wan2.1-1.3B~\cite{wan2025}, with resolution of 720P and supporting maximum 129 frames. It can be adapted for global stylization, video relight, weather/season editing task. Specifically, we use LLM to generate the prompts and ask image editor~\cite{flux2_klein, nano_banana_pro_google_2025} to edit the first frame, then propagate the change with our global editor to the whole video. For Video Recam, we employ Recam-Master~\cite{recammaster_bai2025recammaster} to generate videos with novel camera viewpoints. We treat the edited videos as targets and pair them with their corresponding originals, resulting in 200K reliable video pairs. For VFX Editing, we use Gen4-Aleph~\cite{gen4_aleph} to generate VFX videos. 

\begin{figure*}[t]
\vspace{-1em}
    \centering
    \includegraphics[width=0.985\linewidth]{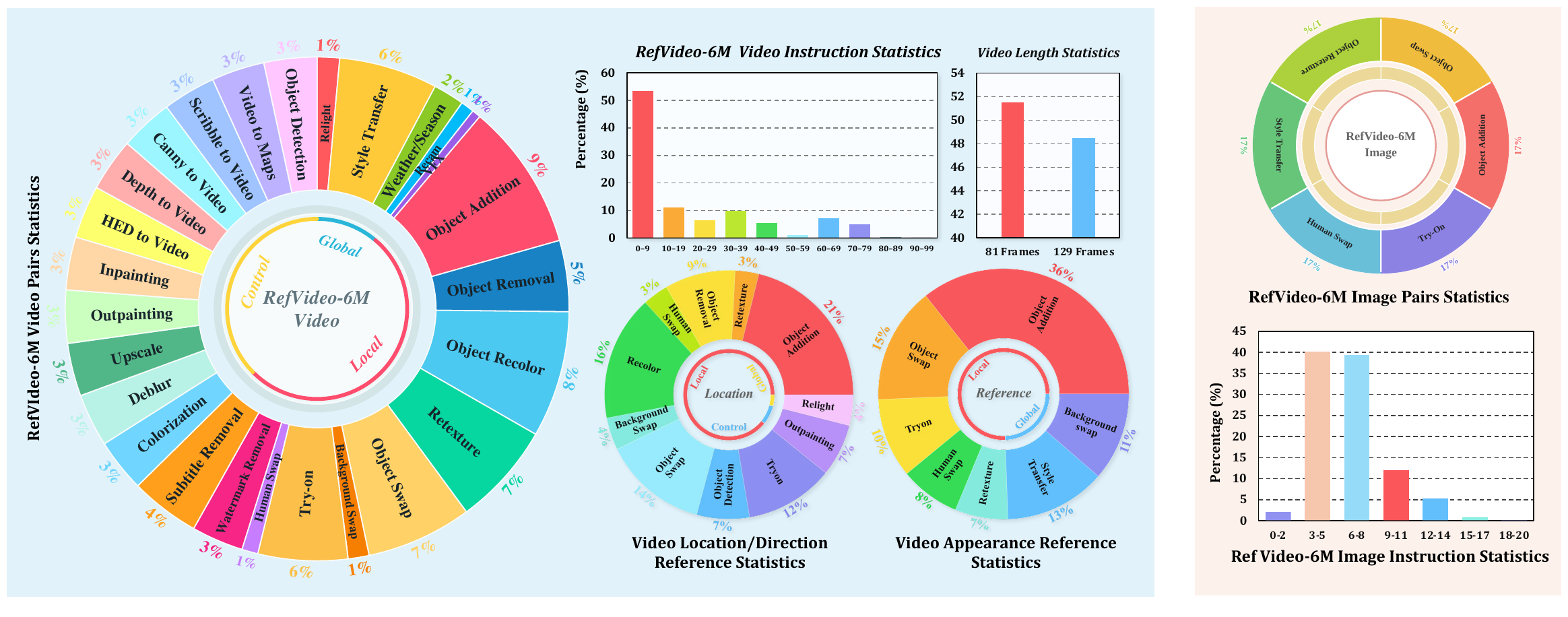}
    \vspace{-1.25em}
    \caption{Statistics of RefVideo-6M. The dataset includes 26 editing tasks and 6 image editing tasks.}
    \vspace{-1.5em}
    \label{fig:statistics}
\end{figure*}

\noindent \textit{\textbf{Local Editing}}. This category consists of 10 tasks: Object Addition, Object Removal, Object Recolor, Object Retexture~\cite{refacade_huang2025refaccadeeditingobjectgiven}, Object Swap, Background Swap, Try-on, Human Swap, Watermark Removal, and Subtitle Removal. \textit{For all tasks, we use an LLM to generate the editing prompts. In addition, we treat the edited videos as sources and the original videos as targets, which helps reduce artifacts in the ground-truth targets.}

For Object Addition, we use MiniMax-Remover~\cite{minimax_remover_zi2025minimaxremovertamingbadnoise} to remove objects from the original videos and use the object-removed videos as sources. For Object Removal, we train an addition editor to synthesize source videos by propagating the first frame edited with FLUX2-Klein~\cite{flux2_klein} to the entire video. For Object Recolor and Object Retexture, we train an outlook editor to modify the color or texture of the target objects while preserving their structure, thereby producing the source videos. For Object Swap, Background Swap, and Try-on, we use VACE-1.3B~\cite{vace_jiang2025vace} to inpaint the original videos with given masks. For Human Swap, we train a human editor to generate human-swapped videos as sources, ensuring action consistency while preserving the background. For Watermark Removal, we synthesize semi-transparent logos and overlay them onto the original videos to construct the sources. For Subtitle Removal, we render captions on the original videos to construct the edited sources.

\noindent \textit{\textbf{Controllable Editing}}. 
This category includes 11 tasks: Colorization, Deblur, Upscale, Outpainting, Inpainting, HED-to-Video, Depth-to-Video, Canny-to-Video, FakeScribble-to-Video, Video-to- (HED / Depth / Canny / FakeScribble), and Object Detection. Except for the last two tasks, the rest tasks take the edited videos as the sources and original videos as the targets. For Outpainting and Inpainting, we corrupt the background or object regions to construct the source videos. For Colorization, Deblur, and Upscale, we convert the original videos to grayscale, blur them, or downsample them, respectively, and use the degraded videos as sources. For HED-to-Video, Depth-to-Video, Canny-to-Video, and FakeScribble-to-Video, we extract the corresponding HED, depth, Canny, and FakeScribble conditions from the original videos and use them as sources. Conversely, for Video-to-HED, Video-to-Depth, Video-to-Canny, Video-to-FakeScribble, and Object Detection, we use the original videos as sources and the extracted video conditions or masks as targets.

\noindent \textit{\textbf{Instruction and Reference Construction}}. The instructions are generated by GPT-5.2~\cite{gpt_5_2_openai_gpt52_2025}. Specifically, we provide the LLM with information about the source and target videos and ask it to generate appropriate editing instructions.

Reference images are divided into two categories: location/direction references and appearance references. For location and direction references, users may provide bounding boxes, circles, and masks as spatial references, as well as direction images to control the light-source direction. We construct such references for the following tasks: Object Addition, Object Removal, Object Recolor, Object Retexture, Object Swap, Background Swap, Try-on, Human Swap, Video Object Detection, Outpainting, and Relight. For appearance references, we generate reference images using current mainstream generative models~\cite{nano_banana_pro_google_2025, flux2_klein} for Object Addition, Object Swap, Object Retexture, Try-on, Human Swap, and Style Transfer. For Background Swap, we instead use MiniMax-Remover~\cite{minimax_remover_zi2025minimaxremovertamingbadnoise} to remove the foreground from the given image and use the resulting background as the reference.

\begin{figure*}[t]
\vspace{-1.75em}
    \centering
    \includegraphics[width=1.0\linewidth]{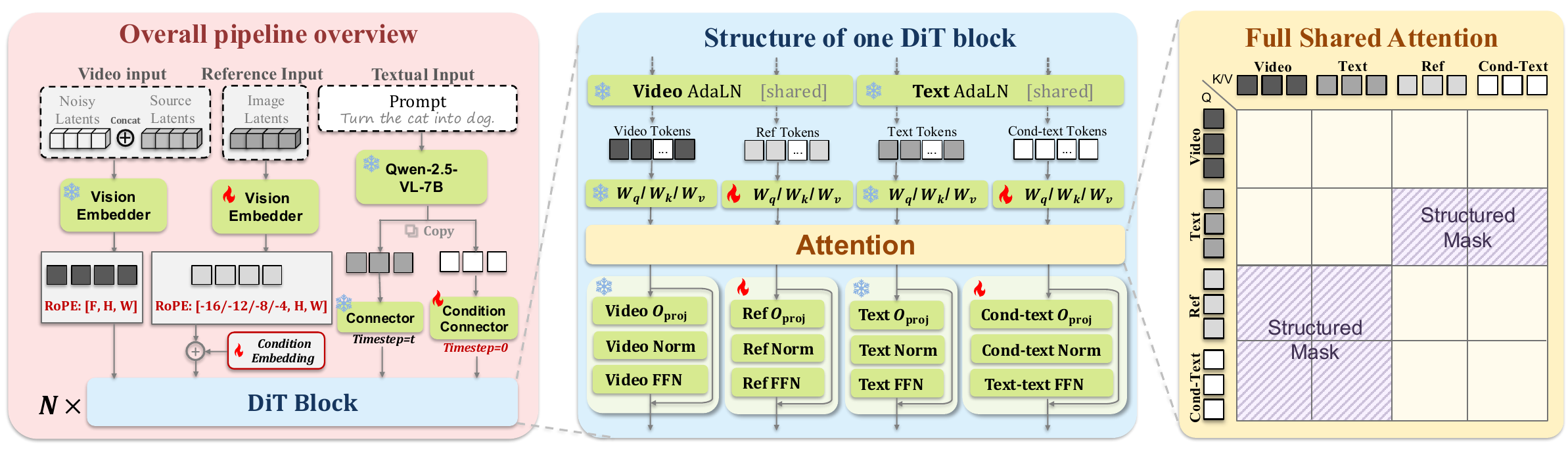}
    \vspace{-2.25em}
    \caption{Overview of RefMoT. RefMoT adopts a Mixture-of-Tokens design to efficiently incorporate reference information, reducing training cost while improving reference-guided editing quality and preserving strong generalization.}
    \vspace{-1.5em}
    \label{fig:arch}
\end{figure*}

\subsubsection{Image Data for Video Editing}

To enhance the video editing dataset, we construct 6 reference-based image editing subsets, including Style Transfer, Object Addition, Object Retexture, Object Swap, Human Swap, and Try-on. Each sample is formulated as a quadruple consisting of a source image, a reference image, a target image, and an editing instruction. Specifically, we first generate the reference image based on the prompt produced by the LLM~\cite{gpt_5_2_openai_gpt52_2025}. Then, conditioned on the reference image, we use an image generator to synthesize the target images. Next, we convert the target images into source images by deleting the reference object for object addition, changing the image style for style transfer, swapping clothing for try-on, modifying object textures for object retexture, swapping humans for human swap, and swapping objects for object swap. Finally, we ask the LLM to generate the corresponding instructions.

\subsubsection{Quality Filtering}
We mainly use GPT-5.2 to filter incorrectly edited videos. For each task, 3 human annotators iteratively refine the quality-check prompt until the LLM can reliably distinguish successful edits from failed ones. Specifically, the LLM filters out edited videos from multiple perspectives. It first deletes videos with visible artifacts to ensure clear visual quality. The LLM then check instruction consistency by verifying whether the source video, target video, and references match the editing instruction. Next, the LLM evaluate motion consistency and identify inconsistent or unnatural motion. Finally, we use LLM detect hallucinated objects or unintended changes outside the specified editing regions. For reference images, we also use the LLM to remove samples with artifacts or mismatched target objects, identities, clothing, textures, or styles. In addition, we use DINO-V3 visual features to filter unchanged edited videos and remove pairs with source-target motion inconsistency.

\subsection{Dataset Statistics}

Our dataset contains approximately 5M video pairs across three categories: global editing, local editing, and controllable editing, with 0.56M, 2.82M, and 1.98M pairs, respectively. It also includes 1M image pairs to further enhance video editing performance. The detailed statistics are shown in Figure \ref{fig:statistics}. Visualizations of our dataset are presented in Figure \ref{fig:teasers}. Our video dataset covers global, local, watermark/subtitle removal, and controllable editing tasks. Global editing includes 335K style-transfer pairs, 45K recam pairs, and 105K weather/season pairs. Local editing includes 429K object-recoloring pairs, 350K retexture pairs, 500K object-addition pairs, 359K object-swap pairs, and 304K virtual try-on pairs. Watermark and subtitle removal contain 175K and 236K pairs, respectively, and each controllable editing task contains approximately 180K pairs. All videos are provided at 720p resolution, including 2.43M videos with 129 frames, 2.83M videos with 81 frames, and VFX videos generated by Gen4-Aleph with 121 frames. The video dataset further includes 5M reference images, consisting of 144K global references, 4.63M local references, and 480K controllable-editing references. Our image dataset consists of 6 reference-based editing tasks: object addition, object retexture, object swap, style transfer, human swap, and try-on. We use Flux2-Klein-9B to construct the image editing data. Each task contains approximately 176K image editing pairs, resulting in a balanced task distribution, and each sample is accompanied by a corresponding reference image.

\renewcommand{\arraystretch}{1.25}
\begin{table*}[!t]
\centering
\setlength{\tabcolsep}{1pt}
\renewcommand{\arraystretch}{1.1}
\resizebox{\textwidth}{!}{
\begin{tabular}{lccccccccccccccc}
\toprule
\multirow{3}{*}{Method}
& \multicolumn{6}{c}{GPT-5.5 Eval}
& \multicolumn{6}{c}{Gemini-3-Pro Eval}
& \multicolumn{2}{c}{Quantitative Results}
& \multirow{3}{*}{User Study} \\
\cmidrule(lr){2-7} \cmidrule(lr){8-13} \cmidrule(lr){14-15}
& \multirow{2}{*}{Bg. Pres.}
& \multicolumn{3}{c}{Instruction Alignment}
& \multirow{2}{*}{Visual Quality}
& \multirow{2}{*}{Overall}
& \multirow{2}{*}{Bg. Pres.}
& \multicolumn{3}{c}{Instruction Alignment}
& \multirow{2}{*}{Visual Quality}
& \multirow{2}{*}{Overall}
& \multirow{2}{*}{CLIPScore} 
& \multirow{2}{*}{Temp-Cons}
& \\
\cmidrule(lr){3-5} \cmidrule(lr){9-11}
&
& Attr. Align.
& Halluc. Supp.
& Overall Instr.
&
&
&
& Attr. Align.
& Halluc. Supp.
& Overall Instr.
&
&
& & & \\
\midrule
InsViE~\cite{wu2025insvie}    & 3.39 & 2.17 & 3.94 & 2.05 & 2.76 & 2.86 & 3.94 & 1.88 & 4.30 & 1.85 & 3.01 & 3.00 & 0.340334 & 0.984086 & 0.3\%\\
InsV2V~\cite{insv2v_cheng2024consistent}    & 3.48 & 2.21 & 4.25 & 2.33 & 3.22 & 3.10 & 4.09 & 1.38 & 4.40 & 1.75 & 3.31 & 2.99 & 0.345979 & 0.978932 & 0.4\%\\
VideoCoF~\cite{videocof_yang2025unifiedvideoeditingtemporal}  & 4.02 & 2.54 & 4.28 & 2.23 & 3.47 & 3.31 & 4.33 & 2.50 & 4.21 & 1.93 & 3.49 & 3.29 & 0.348055 & 0.973684 & 0.6\%\\
ICVE~\cite{icve_liao2025incontextlearningunpairedclips}      & 4.02 & 2.92 & 4.56 & 2.77 & 3.60 & 3.57 & 4.52 & 2.46 & 4.55 & 2.54 & 3.71 & 3.56 & 0.349055 & 0.983105 & 0.3\%\\
ReCo~\cite{reco_zhang2025regionconstraintincontextgenerationinstructional}      & 3.65 & 3.54 & 4.50 & 3.56 & 3.16 & 3.68 & 4.32 & 3.29 & 4.20 & 3.38 & 3.03 & 3.64 & 0.359355 & 0.978573 & 0.3\%\\
Ditto~\cite{ditto_bai2025scalinginstructionbasedvideoediting}     & 3.94 & 3.54 & 4.21 & 3.13 & 3.73 & 3.71 & 4.33 & 3.12 & 4.34 & 2.94 & 3.86 & 3.72 & 0.352119 & 0.981422 & 0.4\%\\
Lucy~\cite{lucy_edit_decart2025lucyedit} & \textcolor{red}{4.18} & 3.29 & 4.46 & 2.93 & 3.72 & 3.72 & 4.55 & 3.21 & 4.51 & 2.75 & 3.69 & 3.74 & 0.358563 & 0.975711 & 0.1\%\\
Senorita~\cite{senorita_zi2025se}  & 3.79 & 4.04 & 4.57 & 4.05 & 3.38 & 3.97 & 4.08 & 4.12 & 4.16 & 4.06 & 3.01 & 3.89 & 0.353698 & 0.973872 & 0.7\%\\
Kiwi~\cite{kiwi-edit}      & 4.02 & 3.83 & 4.70 & 3.72 & \textcolor{blue}{3.77} & 4.01 & 4.50 & 3.88 & 4.40 & 3.60 & 3.64 & 4.00 & 0.355728 & 0.978659 & 6.4\%\\
Capybara~\cite{capybara_2026raocapybara}  & \textcolor{red}{4.18} & 3.96 & 4.59 & 3.46 & \textcolor{red}{3.78} & 3.99 & 4.71 & 3.83 & 4.53 & 3.37 & 3.72 & 4.08 & 0.350744 & 0.984005 & 6.0\%\\
OmniVideo~\cite{omnivideo2_yang2026omnivideo2scalingmllmconditioned} & 3.89 & \textcolor{red}{4.54} & 4.72 & 4.19 & 3.72 & \textcolor{blue}{4.21} & 4.29 & 4.42 & 4.48 & 4.11 & 3.66 & 4.19 & 0.359752 & 0.980869 & 7.7\%\\
UniVideo~\cite{univideo_wei2026univideounifiedunderstandinggeneration}  & 4.08 & \textcolor{blue}{4.50} & 4.66 & 3.87 & 3.68 & 4.16 & \textcolor{blue}{4.73} & 4.38 & \textcolor{blue}{4.61} & 3.81 & 3.68 & 4.24 & 0.356920 & 0.984486 & 6.2\%\\
VINO~\cite{vino_chen2026vino}      & 3.88 & 4.42 & \textcolor{blue}{4.75} & \textcolor{blue}{4.30} & 3.58 & 4.18 & 4.35 & \textcolor{red}{4.58} & 4.51 & \textcolor{blue}{4.31} & 3.64 & \textcolor{blue}{4.28} & \textcolor{blue}{0.364205} & 0.982608 & \textcolor{blue}{7.9\%}\\ 
Bernini~\cite{team2026bernini}      & 3.95 & 4.33 & 4.13 & \textcolor{red}{4.60} & 3.66 & 4.13 & 4.07 & 3.83 & 3.73 & 4.15 & \textcolor{blue}{4.01} & 3.96 & 0.364262 & \textcolor{blue}{0.984766} & 7.1\%\\ \hline
Ours      & \textcolor{blue}{4.12} & 4.25 & \textcolor{red}{4.76} & 4.25 & 3.74 & \textcolor{red}{4.22} & \textcolor{red}{4.83} & \textcolor{blue}{4.50} & \textcolor{red}{4.84} & \textcolor{red}{4.40} & \textcolor{red}{4.25} & \textcolor{red}{4.56} & \textcolor{red}{0.364589} & \textcolor{red}{0.985401} & \textcolor{red}{54.7\%}\\
\bottomrule
\end{tabular}
}
\vspace{-0.5em}
\caption{Evaluation results on instruction-based editing. The best results are marked in red, and the second-best results are marked in blue. Bg. Pres. denotes Background Preservation, Attr. Align. denotes Attribute Alignment, Halluc. Supp. denotes Hallucination Suppression, and Overall Instr. denotes Overall Instruction Alignment.}
\vspace{-1em}
\label{tab:gpt55_eval_ins}
\end{table*}

\begin{figure*}[!htp]
    \centering
    \centering
    \includegraphics[width=\textwidth]{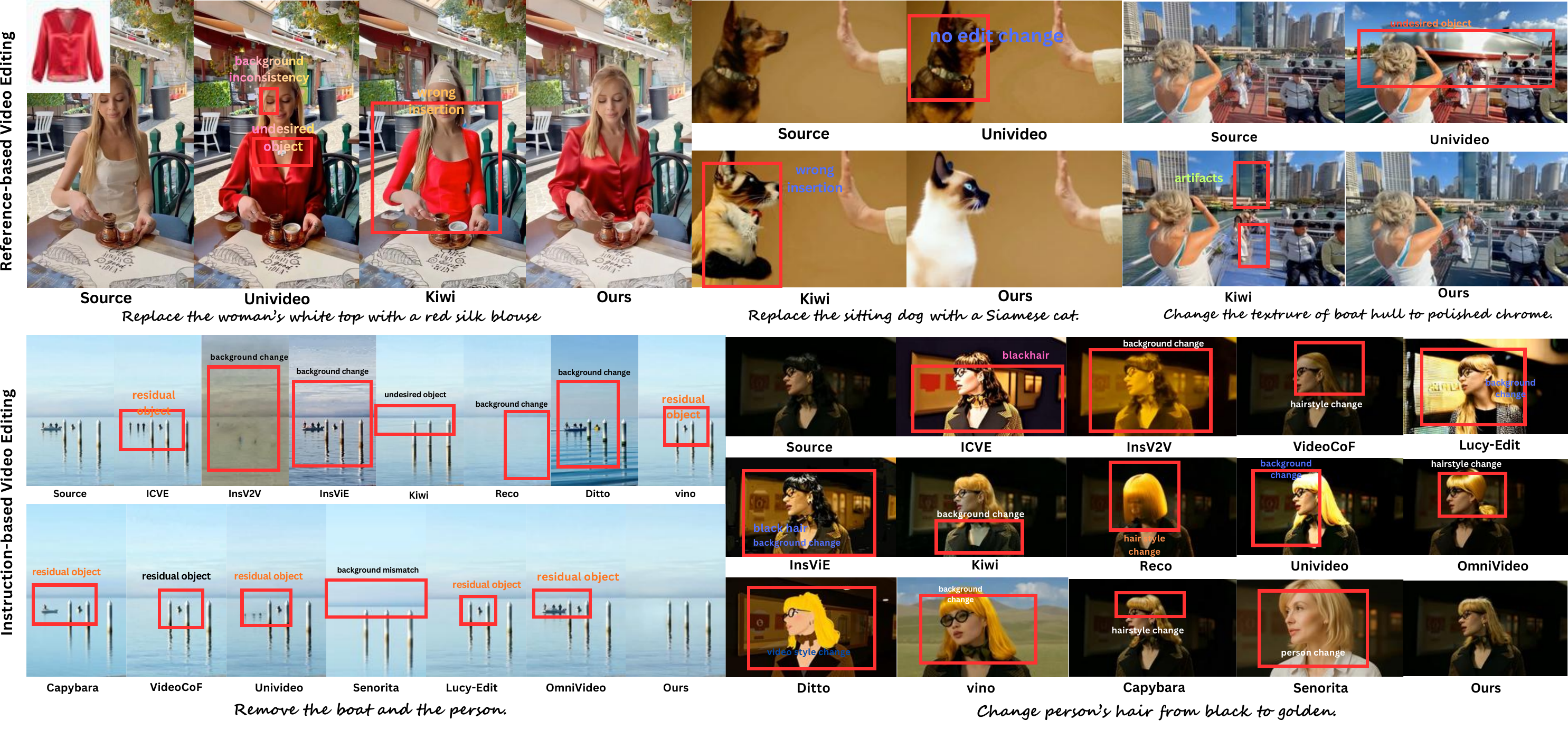}
    \vspace{-1.75em}
    \caption{Visualization of our model and prior methods for instruction-based and reference-based editing.}
    \label{fig:method_visualization}
    \vspace{-1em}
\end{figure*}

\renewcommand{\arraystretch}{1.05}
\begin{table*}[t]
\centering
\setlength{\tabcolsep}{3pt}
\renewcommand{\arraystretch}{1.1}
\resizebox{\textwidth}{!}{
 \begin{tabular}{lccccccccccccccc}
   \toprule
   \multirow{3}{*}{Method}
   & \multicolumn{6}{c}{GPT-5.5 Eval}
   & \multicolumn{6}{c}{Gemini-3-Pro Eval}
   & \multicolumn{2}{c}{Quantitative Results}
   & \multirow{3}{*}{User Study} \\
   \cmidrule(lr){2-7} \cmidrule(lr){8-13} \cmidrule(lr){14-15}
   & \multirow{2}{*}{Bg. Pres.}
   & \multicolumn{3}{c}{Instruction Alignment}
   & \multirow{2}{*}{Visual Quality}
   & \multirow{2}{*}{Overall}
   & \multirow{2}{*}{Bg. Pres.}
   & \multicolumn{3}{c}{Instruction Alignment}
   & \multirow{2}{*}{Visual Quality}
   & \multirow{2}{*}{Overall}
   & \multirow{2}{*}{CLIPScore}
   & \multirow{2}{*}{Temp-Cons}
   & \\
   \cmidrule(lr){3-5} \cmidrule(lr){9-11}
   &
   & Attr. Align.
   & Halluc. Supp.
   & Overall Instr.
   &
   &
   &
   & Attr. Align.
   & Halluc. Supp.
   & Overall Instr.
   &
   &
   & & & \\
   \midrule
     Kiwi~\cite{kiwi-edit}
   & \textcolor{blue}{3.88} & 2.96 & 4.09 & 3.53 & 3.29 & 3.55
   & 4.07 & 2.35 & 3.88 & 2.99 & 2.98 & 3.25
   & 0.346991 & 0.968535 & 15.1\% \\

   UniVideo~\cite{univideo_wei2026univideounifiedunderstandinggeneration}
   & 3.81 & \textcolor{blue}{3.51} & 4.18 & 3.68
   & \textcolor{blue}{3.44} & \textcolor{blue}{3.73}
   & \textcolor{blue}{4.46} & \textcolor{red}{3.93} & \textcolor{blue}{4.22}
   & 3.67 & 3.60 & \textcolor{blue}{3.98}
   & \textcolor{blue}{0.351239} & \textcolor{blue}{0.971174} & 14.5\% \\

   Bernini~\cite{team2026bernini}
   & 3.80 & 3.48 & \textcolor{blue}{4.21} & \textcolor{blue}{3.76}
   & 3.27 & 3.70
   & 4.15 & 3.69 & 3.98
   & \textcolor{blue}{3.73} & \textcolor{red}{3.84} & 3.88
   & 0.350255 & 0.971026 & \textcolor{blue}{17.9\%} \\
   \midrule
   Ours
   & \textcolor{red}{4.22} & \textcolor{red}{3.62} & \textcolor{red}{4.59}
   & \textcolor{red}{4.13} & \textcolor{red}{3.60} & \textcolor{red}{4.03}
   & \textcolor{red}{4.70} & \textcolor{blue}{3.86} & \textcolor{red}{4.51}
   & \textcolor{red}{4.02} & \textcolor{blue}{3.68} & \textcolor{red}{4.16}
   & \textcolor{red}{0.354365} & \textcolor{red}{0.973043}
   & \textcolor{red}{52.5\%} \\
   \bottomrule
   \end{tabular}
}
\vspace{-0.5em}
\caption{Evaluation results on reference-based editing. The abbreviations remain consistent with those in the previous table.}
\label{tab:gpt55_eval_ref}
\vspace{-1em}
\end{table*}

\subsection{Model Architecture and Training}

We train a vanilla instruction-based video editor based on HunyuanVideo1.5~\cite{hunyuanvideo_1_5_wu2025hunyuanvideo15technicalreport}, by concatenating the condition and noisy latents in channel wise. We then propose RefMoT, which injects reference information into the model to better control the editing results.

\subsubsection{Stage-1: Instruction-based Video Editing Model} We train the editing model on instruction-based data using a vanilla video editing architecture~\cite{instructpix2pix_brooks2023instructpix2pix}, where the condition latents and noisy latents are concatenated along the channel dimension. We further modify the system prompt to better adapt it to video editing instructions.

\subsubsection{Stage-2: Reference-based Video Editing Model}
Current reference-based video editing models typically concatenate reference tokens with editing tokens along the token dimension and jointly train on instruction-based and reference-based data. However, this strategy does not directly fine-tune from a pretrained editing model, leading to slower convergence. Moreover, joint training significantly increases the computational cost, while repeatedly training on already converged instruction-based data is inefficient.

To address these issues, we introduce a reference-specific architecture, termed \textbf{\textit{RefMoT}}. Instead of updating the entire model, we freeze the main branch and introduce trainable MoT linear layers dedicated to processing reference tokens. Vision tokens and editing text tokens are still processed by the frozen projections learned in Stage-1, whereas reference tokens are routed through the newly introduced RefMoT layers. Importantly, vision tokens and reference tokens share the same frozen AdaLN layers, which keeps them in a unified feature space and prevents instability caused by learning separate normalization spaces for different token types. The detailed architecture is shown in Figure \ref{fig:arch}. This design preserves the instruction-based editing ability learned in Stage 1, reduces Stage-2 training cost by about 50\% by requiring only reference-based data, and lets the frozen main branch handle instruction understanding and edit localization while the trainable RefMoT layers encode reference tokens as visual guidance.

\renewcommand{\arraystretch}{1.25}
\begin{table*}[!t]
\centering
\setlength{\tabcolsep}{6.0pt}
\renewcommand{\arraystretch}{1.1}
\resizebox{\textwidth}{!}{
\begin{tabular}{llcccccccccccccc}
\toprule
\multirow{3}{*}{Type} & \multirow{3}{*}{Method}
& \multicolumn{6}{c}{GPT-5.5 Eval}
& \multicolumn{6}{c}{Gemini-3-Pro Eval}
& \multicolumn{2}{c}{Quantitative Results} \\
\cmidrule(lr){3-8} \cmidrule(lr){9-14} \cmidrule(lr){15-16}
& & \multirow{2}{*}{Bg. Pres.} & \multicolumn{3}{c}{Instruction Alignment} & \multirow{2}{*}{Visual Quality} & \multirow{2}{*}{Overall}
& \multirow{2}{*}{Bg. Pres.} & \multicolumn{3}{c}{Instruction Alignment} & \multirow{2}{*}{Visual Quality} & \multirow{2}{*}{Overall}
& \multirow{2}{*}{CLIPScore} & \multirow{2}{*}{Temp-Cons} \\
\cmidrule(lr){4-6} \cmidrule(lr){10-12}
& & & Attr. Align. & Halluc. Supp. & Overall Instr. & & & & Attr. Align. & Halluc. Supp. & Overall Instr. & & & & \\
\midrule
\multirow{7}{*}{Arch}
& w/o RefMoT    & 3.95 & 3.38 & 4.15 & 3.79 & 3.05 & 3.66 & 4.10 & 3.58 & 3.74 & 3.51 & 2.73 & 3.53 & 0.357992 & 0.963980 \\
& 1/4 RefMoT    & 4.09 & 2.47 & 4.44 & 3.20 & 3.36 & 3.51 & 4.54 & 2.22 & 4.29 & 2.68 & 3.35 & 3.42 & 0.346781 & 0.975295 \\
& 1/3 RefMoT    & 4.06 & 2.96 & \textcolor{blue}{4.58} & 3.48 & 3.45 & 3.71 & 4.42 & 2.81 & 4.31 & 3.08 & 3.23 & 3.57 & 0.346675 & 0.974725 \\
& 1/2 RefMoT    & 4.06 & 3.09 & 4.47 & 3.65 & 3.45 & 3.75 & 4.37 & 2.99 & \textcolor{blue}{4.34} & 3.32 & 3.23 & 3.65 & 0.349798 & 0.974311 \\
& 3/4 RefMoT    & \textcolor{blue}{4.09} & \textcolor{blue}{3.49} & \textcolor{blue}{4.58} & \textcolor{blue}{4.07} & \textcolor{blue}{3.63} & \textcolor{blue}{3.97} & \textcolor{blue}{4.60} & 3.59 & 4.60 & \textcolor{blue}{3.91} & \textcolor{blue}{3.69} & \textcolor{blue}{4.08} & \textcolor{blue}{0.357294} & \textcolor{blue}{0.974893} \\
& Unfrozen Main & 3.69 & 2.96 & 4.11 & 3.37 & 2.94 & 3.41 & 4.15 & 2.70 & 3.84 & 2.92 & 2.61 & 3.24 & 0.352093 & 0.964379 \\
\cmidrule(lr){2-16}
& Ours          & \textcolor{red}{4.12} & \textcolor{red}{4.25} & \textcolor{red}{4.76} & \textcolor{red}{4.25} & \textcolor{red}{3.74} & \textcolor{red}{4.22} & \textcolor{red}{4.83} & \textcolor{red}{4.50} & \textcolor{red}{4.84} & \textcolor{red}{4.40} & \textcolor{red}{4.25} & \textcolor{red}{4.56} & \textcolor{red}{0.364589} & \textcolor{red}{0.985401} \\
\midrule
\multirow{7}{*}{Dataset}
& INSV2V   & 3.47 & 1.75 & 4.09 & 1.59 & 2.75 & 2.73 & 3.48 & 1.38 & 3.74 & 1.17 & 2.50 & 2.45 & 0.341445 & 0.961964 \\
& Senorita & 3.24 & 1.92 & 3.83 & 2.01 & 2.44 & 2.69 & 3.38 & 1.67 & 3.60 & 1.60 & 2.45 & 2.54 & 0.341658 & 0.976656 \\
& InsViE   & \textcolor{blue}{4.11} & 1.33 & \textcolor{blue}{4.23} & 1.51 & 3.41 & 2.92 & 4.15 & 1.04 & \textcolor{blue}{4.32} & 1.26 & 3.25 & 2.80 & \textcolor{blue}{0.349018} & 0.977576 \\
& ReCo     & 3.68 & \textcolor{blue}{2.75} & 4.00 & 2.12 & 2.97 & 3.10 & 3.95 & 1.67 & 3.32 & 1.61 & 2.63 & 2.64 & 0.341157 & 0.984208 \\
& OpenVE   & 4.08 & 2.46 & 4.16 & 2.04 & \textcolor{blue}{3.54} & 3.25 & 4.15 & 1.88 & 4.19 & 1.73 & \textcolor{blue}{3.52} & 3.09 & 0.346125 & 0.983352 \\
& Ditto    & 3.95 & 2.71 & 4.08 & \textcolor{blue}{2.42} & 3.45 & \textcolor{blue}{3.32} & \textcolor{blue}{4.35} & \textcolor{blue}{2.12} & 4.11 & \textcolor{blue}{2.16} & 3.29 & \textcolor{blue}{3.21} & 0.346923 & \textcolor{blue}{0.984306} \\
\cmidrule(lr){2-16}
& Ours     & \textcolor{red}{4.15} & \textcolor{red}{3.29} & \textcolor{red}{4.62} & \textcolor{red}{3.68} & \textcolor{red}{3.63} & \textcolor{red}{3.88} & \textcolor{red}{4.56} & \textcolor{red}{3.46} & \textcolor{red}{4.44} & \textcolor{red}{3.48} & \textcolor{red}{3.56} & \textcolor{red}{3.90} & \textcolor{red}{0.352300} & \textcolor{red}{0.984560} \\
\bottomrule
\end{tabular}
}
\vspace{-0.65em}
\caption{Ablation results on different model architectures and datasets. The abbreviations remain consistent with those in the previous table.}
\label{tab:gpt55_eval_dataset_refMoT}
\vspace{-1.75em}
\end{table*}

\section{Experiments}
\subsection{Training And Inference Details}

\textbf{\textit{Training Details}}. We train our editing model based on HunyuanVideo1.5 using AdamW~\cite{adamw_loshchilov2017decoupled} with a learning rate of $1\times10^{-5}$ and batch size 64. Videos are sampled at $81\times480\times832$. To reduce memory cost, we update only one-third of the DiT layers and use FSDP2~\cite{pytorch_fsdp2_tutorial} for parameter and gradient sharding. The Stage-2 model, including expert parameters, is initialized from the trained Stage-1 model, and is further trained for one epoch on reference-based image and video editing data, with only expert parameters updated. Other hyperparameters follow Stage-1. More details can be found in the source code we uploaded to the \emph{OpenReview} platform.

\noindent \textbf{\textit{Inference Details}}. All baseline models, except InsViE and Senorita, are evaluated together with our model at 480p resolution with 81 frames, while InsViE and Senorita generate 33-frame videos at the same resolution. For instruction-based editing, our model performs inference in approximately 2 minutes on a single H100 GPU, using 32GB of GPU memory with 25 inference steps and a guidance scale of 4.5. For reference-based editing, our model achieves similar inference time with KV-cache acceleration, using 46GB of GPU memory under the same step and guidance settings.

\subsection{Benchmark Establishment}
To avoid potential overlap with the pretraining data of existing generative models and with our own dataset, we collect videos uploaded to Pexels.com~\cite{pexels} within the past three months and ensure that they are not included in RefVideo-6M. We use 100 videos for instruction-based editing and 100 videos for reference-based editing. For the reference-based setting, we provide diverse reference inputs, including bounding boxes, circle-shaped references, object references, and texture images. We use two mainstream LLMs to evaluate the editing results. The consistency validation between LLM-as-judge and human quality annotations is included in the Appendix.

\subsection{Experimental Results}

\textbf{\textit{Instruction-based Editing}}.
Table \ref{tab:gpt55_eval_ins} reports the quantitative results on instruction-based video editing. Our method achieves the best overall performance under both GPT-5.5 and Gemini-3-Pro evaluations. In the GPT-5.5 Eval, it obtains the highest Overall Score of 4.22 and the best Hallucination Suppression score of 4.76, while maintaining competitive Background Preservation, Attribute Alignment, Overall Instruction Alignment, and Visual Quality scores of 4.12, 4.25, 4.25, and 3.74, respectively. Compared with strong baselines such as OmniVideo, UniVideo, and VINO, our method better avoids undesired visual content while preserving strong instruction-following ability. In the Gemini-3-Pro Eval, our advantage is more pronounced, ranking first in Background Preservation, Hallucination Suppression, Overall Instruction Alignment, Visual Quality, and Overall Score with scores of 4.83, 4.84, 4.40, 4.25, and 4.56, respectively. It improves the Overall Score from the second-best 4.28 to 4.56. Moreover, our method achieves the best CLIPScore and Temporal consistency, and User Study preference of 54.7\%, demonstrating stronger semantic alignment, temporal stability, and user preference.

\noindent \textbf{\textit{Reference-based Editing}}. 
Table~\ref{tab:gpt55_eval_ref} compares different methods on reference-based editing. Since only a few existing methods support this setting, we compare our model with representative baselines, including Kiwi, UniVideo and Bernini. In the GPT-5.5 Eval, our method achieves the best performance across almost all metrics. Compared with UniVideo, our model improves the Overall score from 3.73 to 4.03, with notable gains in Hallucination Suppression and Overall Instruction Alignment, increasing from 4.18 to 4.59 and from 3.68 to 4.13, respectively. This shows that our model better leverages visual references while preserving editing semantics. Under Gemini-3-Pro Eval, our method again achieves the best Overall Score, improving it from 3.98 to 4.16. It also obtains the best scores in Background Preservation, Hallucination Suppression, and Overall Instruction Alignment. Although UniVideo slightly outperforms ours model in Attribute Alignment, our model achieves stronger instruction following and higher visual quality, leading to better overall performance. These results validate the effectiveness of our reference-guided training design and demonstrate strong generalization to both instruction-based and reference-based editing.

\noindent \textbf{\textit{User Study}}. We conduct a user study through an anonymous website, where participants compare videos generated by all methods, including both instruction-based and reference-based approaches. A total of 33 users provided valid responses. The study includes 12 questions for each of instruction-based and reference-based editing. Each video is evaluated with one to four multiple-choice questions covering editing correctness, instruction following, reference consistency, and visual quality. As shown in Tables~\ref{tab:gpt55_eval_ins} and~\ref{tab:gpt55_eval_ref}, our method achieves the highest preference in both settings, with 54.7\% on instruction-based editing and 52.5\% on reference-based editing, far exceeding the second-best results of 7.9\% and 17.9\%. These results show strong alignment with human judgments. More details can be found in the Appendix.

\subsection{Ablation Study}
Our ablation study is organized into two parts. First, we train editors on different datasets using the Stage-1 architecture to evaluate the effectiveness of our data design. For simplicity, we randomly sample up to 200K videos from each dataset. As shown in Table \ref{tab:gpt55_eval_dataset_refMoT}, our dataset achieves the best overall scores under evaluation by both GPT-5.5 and Gemini-3-Pro, improving over the strongest previous dataset from 3.32 to 3.88 and from 3.21 to 3.90, respectively, with especially large gains in overall instruction alignment from 2.42 to 3.68 and from 2.16 to 3.48. Second, we ablate the key modifications in our RefMoT architecture to verify their contributions. Compared with the strongest architectural variant, our full model further improves the overall score from 3.97 to 4.22 on GPT-5.5 and from 4.08 to 4.56 on Gemini-3-Pro, demonstrating that both our dataset construction and architectural improvements consistently enhance performance. The quantitative results also show the same trend as the LLM evaluation. More details are provided in the Appendix.

\section{Conclusion}

In this paper, we introduce RefVideo-6M, a large-scale, reliable, reference-based dataset for video editing. It contains 5M video editing pairs and 1M image editing pairs to improve video editing models. The video subset is built by 12 editing experts, covering 26 editing tasks in 720p resolution with 81 to 129 frames. For most tasks, we use the edited video as the source and the original high-quality video as the target, reducing artifacts from editing pipelines. The image subset provides additional supervision with 6 reference-based editing tasks. RefVideo-6M supports 10 types of references, allowing users to point out the editing location and appearance. Experiments show that models trained on RefVideo-6M achieve SOTA performance, and our proposed RefMoT further improves results.

\bibliographystyle{reference}
\bibliography{reference}

\clearpage
\appendix

\begin{figure*}
    \centering
    \vspace{-1.95em}    \includegraphics[width=0.92\linewidth]{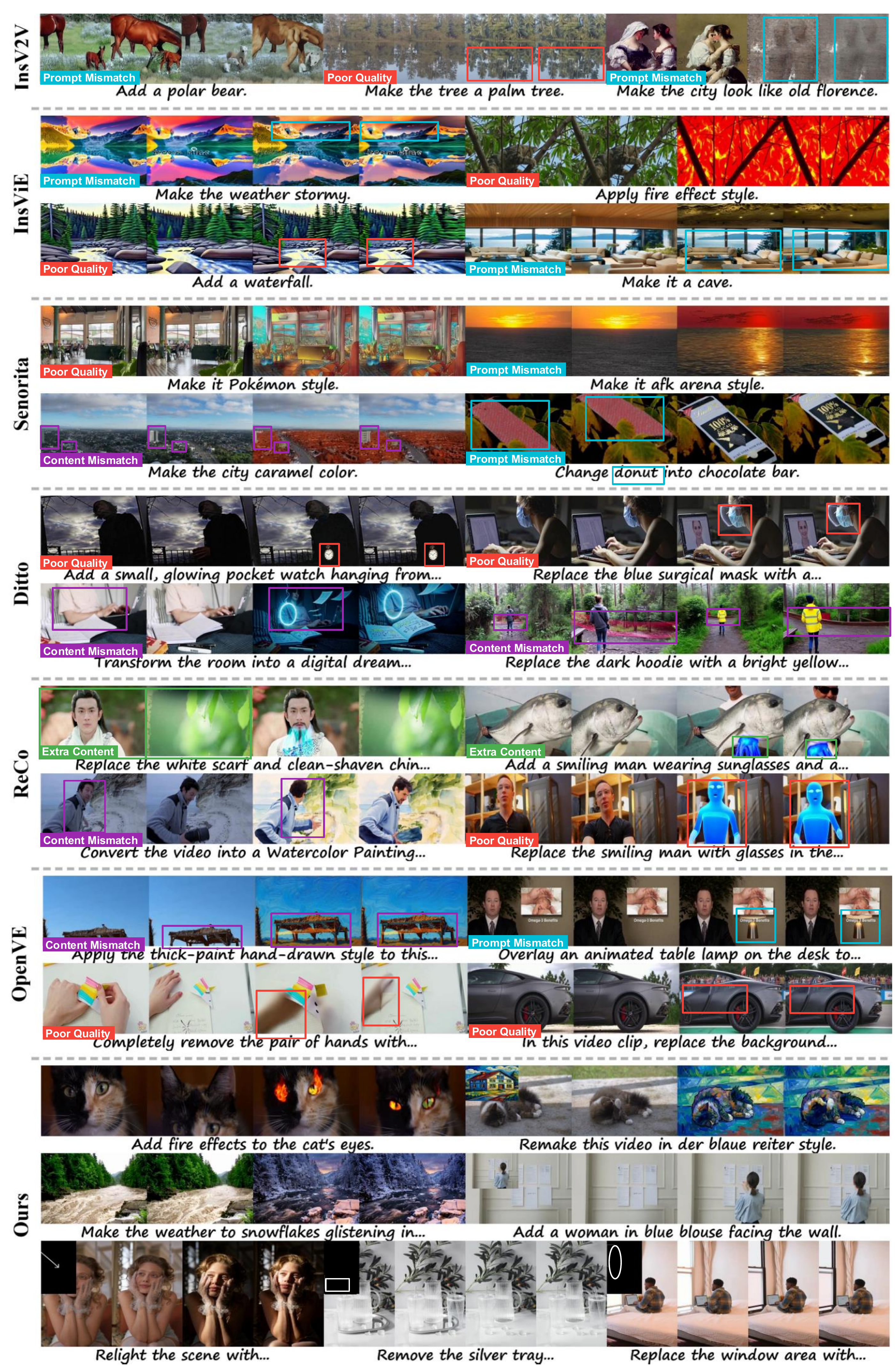}
    \caption{Visualization of baseline datasets and our RefVideo-6M.}
    \label{fig:baseline_refvideo5m_comparison}
\end{figure*}

\section{Training Details of RefMoT}
We train our video editing model in two stages. In Stage 1, the model learns the fundamental source-to-target editing behavior from paired video editing data. Each training sample contains a source video, a target edited video, and a natural-language editing instruction. The source and target videos are resized according to their aspect ratio to either $480{\times}832$ or $832{\times}480$, and each sample is represented with 81 frames. Both videos are encoded into the latent space of a pretrained video VAE. During training, the noisy target latent is concatenated with the source video latent, enabling the transformer to learn how the input video should be modified according to the text instruction.

In Stage 2, we extend the model to support reference-conditioned editing. In addition to the source video and text prompt, the model can take several visual condition images, including reference, bounding-box, circle, and relighting images. These condition images are repeated across the temporal dimension, encoded by the same VAE, and injected into the transformer as auxiliary condition streams. This stage improves the model's ability to follow visual references while preserving the editing capability learned in Stage 1.

Both stages adopt a flow-matching training objective. We randomly sample diffusion timesteps, add noise to the target video latents, and train the transformer to predict the corresponding velocity target. The source video latent serves as the main visual condition, while the editing instruction is encoded by a frozen multimodal text encoder. The VAE and text encoder remain frozen throughout training, and only the transformer parameters are optimized.

The two-stage design separates general editing alignment from reference-conditioned adaptation. Stage 1 aligns the video generation backbone with instruction-based editing, while Stage 2 teaches the model to incorporate additional visual conditions. This design makes the training process more stable and modular, and allows the model to support both text-guided video editing and reference-guided video editing.

We train the model with AdamW optimization, a constant learning-rate scheduler, and a short warm-up period. Mixed-precision training is used to reduce memory consumption and improve efficiency. During training, we periodically run validation sampling with different classifier-free guidance scales to monitor editing quality and guidance sensitivity.

\begin{table*}[t]
\centering
\resizebox{\textwidth}{!}{
\begin{tabular}{llcc}
\toprule
\textbf{Type} & \textbf{Details} & \textbf{Stage 1} & \textbf{Stage 2} \\
\midrule
\multirow{15}{*}{Training}
 & Base model & HunyuanVideo1.5 & Stage-1 model \\
 & Learning rate & $1 \times 10^{-5}$ & $1 \times 10^{-5}$ \\
 & Optimizer & AdamW & AdamW \\
 & AdamW $\beta_1$ & 0.9 & 0.9 \\
 & AdamW $\beta_2$ & 0.99 & 0.99 \\
 & Weight decay & $1 \times 10^{-4}$ & $1 \times 10^{-4}$ \\
 & Training batch size & 64 & 64 \\
 & Training epochs & 1 & 1 \\
 & Training video size & $81{\times}480{\times}832$ & $81{\times}480{\times}832$ \\
 & Training precision & Mixed precision & Mixed precision \\
 & Updated parameters & One-third of DiT layers & Expert parameters only \\
 & Parameter sharding & FSDP2 & FSDP2 \\
 & Training data & Video editing data & Reference-based image and video editing data \\
 & LR scheduler & Constant & Constant \\
 & Diffusion timestep shift & 5.0 & 5.0 \\
\midrule
\multirow{6}{*}{Inference}
 & Inference steps & 25 & 25 \\
 & CFG scale & 4.5 & 4.5 \\
 & Inference precision & fp16 & fp16 \\
 & Inference time & $\sim$2 min on H100 GPU & $\sim$2.5 min on H100 GPU \\
 & Generation resolution & $480{\times}832$ or $832{\times}480$ & $480{\times}832$ or $832{\times}480$ \\
 & Generation frames & 81 & 81 \\
\bottomrule
\end{tabular}%
}
\caption{Training and inference details across different stages. The Stage-2 inference time is measured with the KV-cache version using \texttt{torch.compile}.}
\end{table*}

\section{Expert Construction Details}

Here, we provide the training details for our trained video editing experts. To construct our dataset, we have global stylizer, object addition expert, local stylizer, human swap expert. Specifically, the global stylizer is designed for Relight, Style Transfer, Weather/Season Editing. The local stylizer is designed for Object Recolor and Object Retexture. 

\subsection{Global Stylizer}

Our global stylizer follows the idea of first-frame propagation~\cite{propgen_liu2024generative}, where the edited first frame serves as the reference and its changes are propagated to the entire video to produce the final edited result. This strategy is conceptually simple and has been adopted by many video editing pipelines, including VACE~\cite{vace_jiang2025vace}. Nevertheless, current open-source editing pipelines mainly rely on extracted conditions to guide the propagation process, which inevitably discards motion cues and fine-grained background details from the source video. Consequently, the resulting videos often suffer from suboptimal motion consistency and detail preservation. To overcome these limitations, we retrain a dedicated editor specifically designed for this propagation-based global editing task.

\textbf{Editing Architecture}. Specifically, we adopt Wan2.1-1.3B~\cite{wan2025} as the base model and adapt it into our editing model. The edited first frame is placed in the first temporal slot as the reference latents, with RoPE indices $[0, H, W, C]$, temporal length $F=1$, and timestep $t=0$. A 16-channel vision embedder, initialized from the Wan2.1-1.3B vision embedder, is used to extract the corresponding reference tokens. We then concatenate the noisy latents and source latents along the channel dimension to obtain a 32-channel representation, referred to as the editing latents. These editing latents are appended after the reference latents and assigned RoPE indices $[1:F+1, H, W, C]$, with timesteps sampled from $[0,1]$. To encode them, we introduce a 64-channel editing embedder, whose first 16 input channels are initialized from the original Wan2.1 vision embedder, while the remaining 16 channels are zero-initialized. All other parameters are initialized from Wan2.1-1.3B.

\textbf{Training Dataset Filtering}. We use the global editing subset of Ditto~\cite{ditto_bai2025scalinginstructionbasedvideoediting} as our training dataset. Ditto is constructed by leveraging VACE~\cite{vace_jiang2025vace}, where conditions extracted from the original videos, such as depth maps, are injected into the editing model through a control branch. Given the edited first frame produced by Qwen-Edit~\cite{qwenedit_wu2025qwenimagetechnicalreport}, this pipeline transforms the original videos into the corresponding target videos. However, since the extracted conditions may fail to preserve fine-grained details from the original videos, the target videos in Ditto can lose such details. As a result, they tend to exhibit overly static behavior and reduced motion dynamics. We observe that training on these videos causes the editing model to generate static editing results that fail to match the motion patterns of the original videos.

To address these issues and further improve model performance, we introduce a set of filtering strategies. We first use DINOv3~\cite{dinov3} to extract frame-level features and compute the similarity between consecutive frames. The similarities are then averaged over the entire video to obtain a video-level similarity score, which serves as an indicator of motion variation: a higher score corresponds to weaker motion. We discard samples where the source video has lower similarity than the target video, as this suggests that the target video becomes relatively more static after editing. In addition, we remove samples where both the source and target videos have similarity scores above 0.99, since these videos contain very limited motion. This filtering process effectively removes relatively static or motion-degraded videos from the training dataset.

\textbf{Training Settings}. We train the model using the AdamW optimizer~\cite{adamw_loshchilov2017decoupled} and unfreeze all parameters. Mixed-precision training is adopted, with FP16 forward computation and FP32 gradients and parameter updates. We set the learning rate to $1\times10^{-5}$, weight decay to $1\times10^{-4}$, and use an effective batch size of 64, achieved with a per-step batch size of 32 and 2-step gradient accumulation. Gradient checkpointing is enabled, and the model is trained for 10 epochs. The textual input is always set to null text, as the first frame is edited by an external image editor and textual guidance is therefore unnecessary for the propagation process. Most training videos contain 81 frames, while some videos contain 101 frames to improve training robustness. All videos are resized to 480P resolution, with portrait videos at $480\times832$ and landscape videos at $832\times480$.

\subsection{Local Stylizer}
The local stylizer is designed to generate video pairs for Object Recolor and Object Retexture. Unlike previous works, such as Ditto~\cite{ditto_bai2025scalinginstructionbasedvideoediting} and OpenVE~\cite{openve3m_he2025openve3mlargescalehighqualitydataset}, which typically use editing experts to produce target videos, we instead apply the editing expert to generate the source videos. In this way, the target videos remain free from editing-induced artifacts and other visual flaws. Accordingly, our local stylizer is designed to alter the original color and texture while preserving the object structure and skeleton.

\textbf{Editing Architecture}.
We build our model upon Wan2.1-1.3B and adopt a ControlNet-style inpainting architecture that jointly incorporates structural conditioning and video inpainting. This design enables the text prompt to control the color and texture of the edited regions while preserving their original structure. The model contains two coupled branches: a control branch with $N$ DiT blocks and an inpainting branch with $M$ DiT blocks, where $M$ is divisible by $N$. All parameters are initialized from Wan2.1-1.3B. The control branch takes noisy latents and condition latents as inputs. The condition latents are extracted from structural video conditions, such as Canny edges or HED maps. Specifically, the noisy latents and condition latents are passed through two separate vision embedders, and their resulting token representations are summed. The fused tokens are then processed by the $N$-block control branch to produce multi-level structural control features. The inpainting branch takes 48-channel edit latents as input, constructed by concatenating noisy latents, masked video latents, and mask latents. Given an input video $x$ and a binary mask $m$, the noisy latents are obtained by adding random noise to $\mathcal{E}(x)$, while the masked video latents and mask latents are computed as $\mathcal{E}((1-m) \odot x)$ and $\mathcal{E}(m)$, respectively, where $\mathcal{E}$ denotes the VAE.

To inject structural guidance, the output hidden state of each DiT block in the inpainting branch is augmented with the corresponding hidden state from the control branch. Since the inpainting branch contains $M$ blocks and the control branch contains $N$ blocks, with $M$ divisible by $N$, we use periodic indexing: the $i$-th inpainting block receives the control feature from the $(i \% N)$-th control block. This allows structural control signals to guide the inpainting process throughout the network while preserving the generative prior inherited from Wan2.1-1.3B.

\textbf{Training Dataset Construction}. We construct our training dataset from videos collected from Pexels.com. For each video, we use Qwen3-VL-8B to extract the object names, which are later used as lightweight textual prompts during training. We then apply Rex-Omni~\cite{rex_omni_jiang2025detectpointprediction} to obtain object detection results and propagate the detected bounding boxes across the entire video to generate object masks. To remove noisy and fragmented mask regions, we further filter the generated masks using connected-component analysis.

\textbf{Training Settings}. During training, structural conditions, including Canny edges and HED maps, are extracted online by the condition extractor. We optimize only the control branch and the vision embedder of the inpainting branch, while keeping the remaining parameters frozen. Since our task mainly focuses on modifying the color and texture of the edited regions rather than changing their structure, all textual inputs are constructed as short prompts using the corresponding object names. Unless otherwise specified, we follow the training details of Global Stylizer.

\subsection{Object Removal}

Existing object removal datasets are typically constructed by applying video inpainting methods or using rendering engines such as UE5 to synthesize paired videos. However, inpainting-based pipelines often fail to faithfully capture object-related side effects, such as shadows, illumination changes, and reflections, while engine-based synthesis requires substantial manual effort and is difficult to scale. To build scalable training pairs that support the removal of both objects and their associated side effects, we instead use an object addition model to insert objects into original videos. The edited videos are used as source videos, and the original videos are used as targets. Since the target videos are captured before object insertion, they are naturally free from the added objects and their corresponding side effects.

Nevertheless, constructing a reliable object addition model is non-trivial. In our experiments, direct video object addition tends to produce artifacts, making the resulting pairs less reliable. To mitigate this issue, we adopt a first-frame-guided video object addition pipeline. Specifically, we first use an image editor to generate an edited first frame with the desired object, and then use this frame to guide the video addition model to insert the object consistently across the video. This design improves the controllability of object insertion and enables scalable construction of high-quality object-removal training pairs.

\textbf{Editing Architecture}. 
It is intuitive to adopt the architecture of the Full-Dit (e.g. the architecture used in global stylizer)~\cite{fulldit_ju2025fullditmultitaskvideogenerative}. However, the editing results is suboptimal. We attribute this limitation to the training-test discrepancy caused by the current image editor. During training, the source and target videos are pixel-aligned, while at test time the edited first frame may contain small spatial shifts or local distortions. Such discrepancies make the model overly sensitive to pixel-level misalignment and prevent it from reliably propagating the edit from the first frame to the entire video.

To mitigate this issue, we design an architecture that can has robustness resist to minor spatial shifts or distortions. We feed the edited first frame latents and noisy latents into the main branch and encode the source video with controlNet branch. Specifically, the controlNet takes the source latents and noisy latents as inputs, processes them with two separate embedders, and fuses the resulting tokens by addition. Its hidden states are then periodically injected into the main branch. This design allows the main branch to maintain its generative flexibility while reference the object in the first edited frame and the background in the source videos.

\textbf{Training Dataset Construction}. We use Minimax-Remover~\cite{minimax_remover_zi2025minimaxremovertamingbadnoise} to construct object-removal training pairs from videos collected from Pexels.com. We first use Qwen3-VL-8B~\cite{Qwen3-VL} to extract candidate object names from each video, and then apply Rex-Omni~\cite{rex_omni_jiang2025detectpointprediction} to detect the corresponding objects. Given the detected instances, we use SAM2 to obtain their instance-level masks. For each video, we randomly select one mask and feed it, together with the original video, into Minimax-Remover to generate an edited video with the selected object removed. During inference, we set the resolution to 480P and use 12 sampling steps. The edited videos are used as source videos, and the original videos are regarded as target videos. Finally, this pipeline produces 600K video pairs for training.

\textbf{Training Settings}.
During training, we update only the ControlNet branch and the embedders of the main branch in the object addition model, and freeze all other parameters. The remaining training configurations are kept the same as those used for training the previous expert models.

\subsection{Human Swap}

Existing object swap editors, such as inpainting-based models including VACE~\cite{vace_jiang2025vace} and VideoPainter~\cite{bian2025videopainter}, are capable of swapping general objects. However, when applied to human swap, they often fail to preserve the pose consistency between the original subject and the edited result. This limitation MoTivates us to develop a dedicated human swap model that explicitly focuses on maintaining human pose during the swapping process.

\textbf{Model Architecture}. 
We adopt an architecture similar to the local stylizer. Unlike the local stylizer, which may rely on multiple control signals, our human swap model uses only the pose map as the control condition. Specifically, the control branch takes both the pose latents and the noisy latents as input, while the main branch only receives the noisy latents. This design allows the model to inject pose guidance through the control branch while preserving the generation capacity of the main diffusion branch.

\textbf{Training Dataset Filtering}. 
The training videos are collected from Pexels.com. We first use Qwen3-VL-8B to filter the videos and ensure that valid human subjects are present. For each selected video, we apply OpenPose~\cite{openpose_cao2019openposerealtimemultiperson2d} to extract the pose map of each person. The foreground masks are obtained using the same pipeline as in our object removal model.

We observe that the inpainting results are highly correlated with the shape of the input mask. For example, when training a vanilla human swap model to replace a woman with a man, directly using the original mask of the woman may cause the model to generate a man with long hair in the edited video. This indicates that the model can unintentionally associate the mask shape with the appearance attributes of the source subject, leading to undesired entanglement between the mask geometry and the inpainted content. To resolve this issue, we introduce two types of masks during training. Specifically, we use the circumscribed rectangular masks of the original human masks for 50\% of the training samples, and use the original human masks for the remaining 50\%. The circumscribed rectangular masks provide a less shape-specific spatial constraint, while the original masks preserve accurate foreground regions. By training with both mask types, the human swap model becomes less sensitive to detailed mask geometry and can better disentangle the mask shape from the generated human appearance.

We use short textual prompts based on the target object category names during training. For human swap, the prompts are constructed using simple descriptions such as the target gender or identity category, e.g., ``a man'' or ``a woman''. This concise prompt design encourages the model to focus on the swapping target while relying on the pose condition to preserve the original human structure and motion.

\textbf{Training Settings}. 
During training, we update only the control branch while keeping the remaining components frozen. We use an effective batch size of 64, which is achieved with a mini-batch size of 32 and gradient accumulation over 2 steps. The model is trained for 1 epoch. All other training configurations, including the optimizer, learning rate schedule, noise sampling strategy, and data preprocessing settings, follow those used in the previous models.

\subsection{Distillation Acceleration}

We adopt DMD2~\cite{dmd2_yin2024improveddistributionmatchingdistillation} to distill the editing model, with the goal of reducing the number of inference sampling steps from 50 to 10. In practical editing scenarios, we observe that optimizing with the DMD2 loss alone leads to fast convergence, but often fails to preserve background consistency. To alleviate this issue, we introduce an additional MSE-based flow-matching regularization term. The final training objective is defined as:
\begin{equation}
    \mathcal{L} = \mathcal{L}_{\mathrm{DMD2}} + \lambda \mathcal{L}_{\mathrm{FM}},
\end{equation}
where
\begin{equation}
    \mathcal{L}_{\mathrm{FM}} =
    \mathbb{E}_{t, x_t, c}
    \left[
    \left\|
    v_{\theta}(x_t, t, c) - v
    \right\|_2^2
    \right].
\end{equation}
Here, $\lambda$ controls the strength of the regularization term, $v_{\theta}$ denotes the velocity predicted by the distilled editing model, and $v$ denotes the target velocity used in flow matching. This auxiliary term encourages the distilled model to better match the flow trajectory, thereby improving structural and background consistency during editing.

\textbf{Training Details}. 
We set the hyperparameter $\lambda$ to 1. The model is trained with an effective batch size of 64, consisting of a mini-batch size of 32 and 2 gradient accumulation steps.  The learning rate is set to $5 \times 10^{-6}$, and the weight decay is set to $1 \times 10^{-4}$. 
Gradient checkpointing is enabled to reduce memory consumption during training.  We train the model at resolutions of $480 \times 832$ and $832 \times 480$, using video clips with 81 or 101 frames.  We adopt mixed-precision training, where the forward pass is performed in FP16, while the backward pass and parameter updates are conducted in FP32 for improved numerical stability. Training is run for a maximum of 600 optimization steps.

\section{Video Dataset Construction Details}
We downloaded videos from the pexel.com with authenticated APIs. We use the Dinov3~\cite{dinov3} to compare different videos and then remove the duplicated videos. We cut the video with a segment consisting of 161 frames.

\subsection{Global Editing Video Pairs Construction Details}

\subsubsection{Relight}\label{relight_construction}

We select the 400K source videos from a pool of filtered videos. For each selected video, we use GPT-5.2 to enumerate relighting directions and generate corresponding relighting prompts, covering attributes such as illumination intensity, color temperature, and shadow direction.

\textbf{Editing Process}. Given these prompts, we edit the first frame using flux2-klein-9b~\cite{flux2_klein} and nano banana-pro~\cite{nano_banana_pro_google_2025}. Specifically, flux2-klein-9b is used to obtain standard relighting results, while nano banana-pro is used to generate more challenging relighting cases, particularly those where flux2-klein-9b tends to fail. We then use the edited first frame as a conditioning signal and apply our distilled global stylizer to propagate the relighting effect from the first frame to the entire video. All videos are generated at 720P resolution with 81 frames, using 10 sampling steps during inference. 

\textbf{Quality Filtering}. After obtaining the edited videos, we further filter them using GPT-5.2. To ensure the reliability of the automatic filtering process, human annotators evaluate the filtering accuracy, and we iteratively refine the filtering instructions until the accuracy reaches a satisfactory level. Finally, we use GPT-5.2 to remove videos that contain visible artifacts or motion inconsistencies.

\textbf{Instruction Generation}. We use the qwen-vl-8b to generate the instruction, directly transforming the prompt for first frame editing to the final editing instruction.

\subsubsection{Style Transfer}

We sample 600K videos from the video pool and use GPT-5.2 to generate more than 200 common editing styles, which are further converted into detailed editing prompts. 

\textbf{Editing Pipeline}. For each video, nano banana-pro is used to edit the first frame according to the prompt, and our distilled global stylizer then propagates the edit to the full video. We perform inference at 720P resolution with 81 frames and 10 sampling steps. 

\textbf{Quality Filtering}. This part is same as the Sec. \ref{relight_construction}.

\textbf{Instruction Generation}. This part is same as the Sec. \ref{relight_construction}.

\textbf{Reference Construction}. We use Flux2-Klein-9B to generate the reference style image. We then employ GPT-5.2 to verify whether the generated reference image accurately reflects the intended style.

\subsubsection{Weather/Season}

We sample 300K source videos from video pool and use GPT-5.2 to enumerate editing factors, including weather, season, and time of day (e.g., morning, noon, and evening). These factors are further converted by GPT-5.2 into detailed editing prompts according to the source videos.

\textbf{Editing Pipeline}. We then edit the first frame using Flux2-klein-9b~\cite{flux2_klein} for simple prompts and nano banana-pro~\cite{nano_banana_pro_google_2025} for complex prompts, where prompt difficulty is manually verified beforehand. The resulting first-frame edit is propagated to the full video using our distilled global stylizer, with 720P and 81 frames per video. The data quality filtering operation is same as the Style Transfer. 

\textbf{Quality Filtering}. This part is same as the Sec. \ref{relight_construction}.

\textbf{Instruction Generation}. This part is same as the Sec. \ref{relight_construction}.

\textbf{Reference Construction}. We use Qwen3-VL-8B to parse the relighting instruction used for first-frame editing and extract the corresponding lighting direction. We then render this direction as an arrow on a blank canvas using OpenCV, where the arrow visually specifies the desired relighting direction. The rendered canvas serves as the reference image.

\subsubsection{Recam}

We first select 400K videos from the video pool and 10 candidate camera trajectories. 

\textbf{Editing Process}. For each video, one trajectory is randomly selected and used by recam-master~\cite{recammaster_bai2025recammaster} to generate a synthetic video at 720P resolution and 81 frames with 50 sampling steps. The synthesized video serves as the source video, and the original video is treated as the target video. We further use GPT-5.2 to estimate the trajectory of the original video, providing trajectory annotations for the target. 

\textbf{Quality Filtering}. This part is same as the Sec. \ref{relight_construction}.

\textbf{Instruction Generation}. We feed the trajectory information detected by gpt5.2 and then use the Qwen3-VL-8B to generation the final instruction.

\subsubsection{VFX Editing}

To construct the video editing dataset, we first randomly sample videos from a large video pool. Due to the high cost of the Runway API, we only use it to generate the necessary number of edited videos. Our experiments in the main body show that 654 video pairs are sufficient for the model to acquire the VFX editing capability.

\textbf{Editing Process}. For each selected video, GPT-5.2 was used to generate 7 categories of visual effects (VFX) editing instructions, aiming to cover the most common VFX editing needs. We then randomly select editing category, and use GPT-5.2 further converted the selected editing type into a detailed text prompt. This prompt was used to call the Runway Gen-4 Aleph API to generate edited videos. Through this pipeline, we obtained 654 successful generated examples. Each final video consists of 121 frames with a resolution of 720P. 

\textbf{Quality Filtering}. This part is same as the Sec. \ref{relight_construction}.

\textbf{Instruction Generation}. We use the same instruction as that adopted for calling the Runway API.

\subsection{Local Editing Video Pairs Construction Details}

\subsubsection{Object Addition}\label{object_addition_construction}

To construct the object removal editing data, we selected 800K triplets of videos, object masks, and object names from the video pool. 

\textbf{Editing Process}. The selected videos and corresponding masks were then fed into MiniMax-Remover~\cite{minimax_remover_zi2025minimaxremovertamingbadnoise} to remove the specified objects. The editing was performed at 720P resolution, with each output video containing 81 frames and using 10 sampling steps. After obtaining the edited videos, we employed GPT-5.2 together with human annotator feedback to refine the filtering instructions. For the final training pairs, we treated the original video as the target video and the object-removed edited video as the source video, forming paired examples for object restoration editing. 

\textbf{Quality Filtering}. This part is same as the Sec. \ref{relight_construction}.

\textbf{Instruction Generation}. We first prompt GPT-5.2 to construct a diverse pool of editing-instruction templates. For each instance, we instantiate a selected template with the added object name and then use Qwen3-VL-8B to polish the instruction for clarity and fluency.

\textbf{Reference Construction}. The reference condition for this task includes three possible forms: bounding-box references, circle references, and object references. We construct bounding-box and circle references using OpenCV with simple geometric transformations. For object-level references, we employ nano-banana-pro to generate the corresponding reference object based on the target video. Finally, GPT-5.2 is used as an automatic verifier to remove failed generations.

\subsubsection{Object Removal}\label{object_removal_construction}

We selected 800K triplets of video, mask, and object name from the video pool. 

\textbf{Editing Process}. For each video, we then randomly selected a suitable object to be added in a way that was compatible with the scene content. Based on the selected video and target object, we generated a text prompt describing the desired object addition. We first used FLUX2-Klein-9B to generate the edited first frame, and then propagated this edited first frame through the object addition model to obtain the full edited video. The generation was performed at 720p resolution, with each video containing 81 frames and using 10 sampling steps. 

\textbf{Quality Filtering}. 
Unlike the filtering procedures used in the aforementioned subsets, the object removal subset requires a dedicated filtering strategy to handle two specific challenges. 
First, we address referential ambiguity caused by the presence of same-category objects in the target video, i.e., the original video after removal. Since such objects may make the removal instruction ambiguous, we use Qwen-VL-8B to verify whether the target video still contains objects belonging to the same category as the removed object. If so, the corresponding sample is discarded. Second, we filter out samples with poor background consistency. We use Rex-Omni~\cite{rex_omni_jiang2025detectpointprediction} to detect the location of the removed object in the source video, and then use DINOv3~\cite{dinov3} to measure the similarity between the surrounding background regions in the source and target videos. We retain only samples whose background similarity exceeds 0.985.

\textbf{Instruction Generation}. This part is same as the Sec. \ref{object_addition_construction}.

\textbf{Reference Construction}. The reference condition for this task includes two possible forms: bounding-box references, circle references, and object references. We construct bounding-box and circle references using OpenCV with simple geometric transformations.

\subsubsection{Object Recolor}

We first asked GPT-5.2 to enumerate 200 color names. We then selected 800K triplets of video, mask, and object name from the video pool. 

\textbf{Editing Process}. For each triplet, a color was randomly sampled from the predefined color list, and a text prompt was generated by applying the sampled color to the target object. We used our local stylizer to modify the object color according to the prompt, producing edited videos at 720p resolution, with 81 frames and 10 sampling steps. This process intentionally alters the original color appearance of the object. For the final paired data, we treated the original video as the target video and the color-edited video as the source video. 

\textbf{Quality Filtering}. Following the object addition pipeline, we further adopt the same GPT-5.2-based filtering procedure with human annotator feedback to assess the quality of the edited videos. Moreover, we employ DINOv3~\cite{dinov3} to compute the similarity between the source and target videos, and discard pairs with excessively high similarity to ensure meaningful visual changes. Moreover, the dramatic change in the background is also not permitted, these samples will be removed from the dataset, too.

\textbf{Instruction Generation}. We use Qwen-VL-8B to generate unambiguous editing instructions for the object recolor task. Specifically, Qwen-VL-8B first determines whether the video contains only one object of the same class as the recolored object. When multiple same-class objects are present, it further identifies the target object by describing its relative position and extracting its color and other discriminative attributes, thereby reducing referential ambiguity. We additionally detect the color of the corresponding object in the target video. The extracted information is then integrated by Qwen-VL-8B to compose the final instruction.

\textbf{Reference Construction}. This process follows the same procedure as described in Sec.~\ref{object_removal_construction}.

\subsubsection{Object Retexture}

For the texture editing data, GPT-5.2 was first asked to enumerate 200 texture names. We then selected 800K triplets of video, mask, and object name from the video pool. 

\textbf{Editing Process}. For each triplet, a texture was randomly sampled from the predefined texture list, and a text prompt was generated by applying the sampled texture to the target object with qwen3-vl-8b. A local stylizer was used to modify the object texture according to the prompt, producing edited videos at 720p resolution, with 81 frames and 10 sampling steps. This process intentionally breaks the original texture appearance of the object. For the final paired data, the original video was treated as the target video, while the texture-edited video was treated as the source video. 

\textbf{Quality Filtering}. The quality filtering process  is same as them in the Sec. \ref{object_addition_construction}.

\textbf{Instruction Generation.}
When the video contains multiple objects of the same category as the edited object, we use Qwen3-VL-8B to localize the edited object and extract its visual attributes, thereby reducing referential ambiguity. We additionally use Qwen-VL-8B to identify the texture of the target object. The extracted object-specific information, together with a randomly selected instruction template, is then fed into Qwen3-VL-8B to produce the final editing instruction.

\textbf{Reference Construction}. The reference condition for this task includes three possible forms: bounding-box references, circle references, and object references. We construct bounding-box and circle references using OpenCV with simple geometric transformations. Since generating high-quality texture references is challenging for Flux2-Klein-9B, we use Nano-Banana-Pro to synthesize the texture reference image.

\subsubsection{Object Swap}\label{object_swap_construction}

For the object swap data, we selected 1.8M triplets consisting of a video, a mask, and an object name from the video pool. 

\textbf{Editing Process}. Because the editing success rate for object swaps was relatively low, a large candidate set was required. For each triplet, we used Qwen3-VL-8B to generate a suitable replacement object. Given that Qwen3-VL-8B was sufficiently capable for this relatively constrained task, we did not use the more expensive GPT-5.2 at this stage. Based on the selected replacement object, we then generated a text prompt and used VACE to perform the object swap, producing edited videos at 720p resolution with 81 frames and 25 sampling steps. For the final paired data, the original video was treated as the target video, while the edited video was treated as the source video. 

\textbf{Quality Filtering}. As in the object addition pipeline, we adopt the same GPT-5.2-based filtering procedure with human annotator feedback, where the filtering instructions are iteratively refined to remove videos with artifacts, poor motion consistency, blur, or other quality issues. We further conduct background and overall similarity comparisons to discard samples with drastic background changes or unchanged foreground regions.

\textbf{Instruction Generation.}
We use Qwen-VL-8B to check whether the source video contains multiple objects of the same category as the edited object. If ambiguity exists, Qwen-VL-8B localizes the edited object and extracts its visual attributes. These details are then combined with a randomly selected GPT-5.2-generated instruction template and fed into Qwen-VL-8B to generate an unambiguous editing instruction.

\textbf{Reference Construction}. This process follows the same procedure as described in Sec.~\ref{object_addition_construction}.

\subsubsection{Background Swap}

For the background swap data, we selected 1M triplets of video, mask, and object name from the video pool. 

\textbf{Editing Process}. As with object swap, the editing success rate for background replacement was relatively low, so a large candidate set was required. For each triplet, we used Qwen3-VL-8B to generate a suitable background. Based on the generated background, we then constructed a text prompt and used VACE~\cite{vace_jiang2025vace} to perform background editing, producing edited videos at 720p resolution with 81 frames and 25 sampling steps. For the final paired data, the original video was treated as the target video, while the edited video was treated as the source video. 

\textbf{Quality Filtering}. As in the object addition pipeline in Sec \ref{object_addition_construction}, we apply the same GPT-5.2-based filtering procedure with human annotator feedback, where the filtering instructions are iteratively refined to remove videos with visual artifacts, poor motion consistency, blur, or other quality issues. In addition, we use DINOv3~\cite{dinov3} to filter out source videos with drastic foreground changes.

\textbf{Instruction Generation}. This type of editing also suffers from object-reference ambiguity. When multiple objects in the source video belong to the same category as the edited object, we use Qwen-VL-8B to localize the edited object and extract its visual attributes. We also use Qwen3-VL-8B to describe the background of the target video. The resulting object-specific and background information is then combined with a randomly selected GPT-5.2-generated instruction template and fed into Qwen-VL-8B to produce a precise and unambiguous editing instruction.

\textbf{Reference Construction}. Circle and bounding-box references are constructed following the same procedure as described in Sec.~\ref{object_addition_construction}. For background reference construction, we use MiniMax-Remover~\cite{minimax_remover_zi2025minimaxremovertamingbadnoise} to remove the target object from the first frame. We then use GPT-5.2 to verify whether the original object has been successfully removed and filter out failed cases.

\subsubsection{Tryon}

For the tryon data, we selected 800K triplets of video, mask, and object name from the video pool. 

\textbf{Editing Process}. We first used Qwen3-VL-8B to verify that the object name referred to a clothing item. For each validated triplet, Qwen3-VL-8B was then used to generate a suitable new garment. Based on the generated clothing item, we constructed a text prompt and used VACE to edit the clothing region, producing edited videos at 720p resolution with 81 frames and 25 sampling steps. For the final paired data, the original video was treated as the target video, while the edited video was treated as the source video. 

\textbf{Quality Filtering}. The quality filtering process is same as the object swap in Sec \ref{object_swap_construction}.

\textbf{Instruction Generation.}
When multiple people appear in the video, we first use Qwen3-VL-8B to determine which person corresponds to the edited clothing. We then use Qwen-VL-8B to extract the person's visual attributes, and use Qwen3-VL-8B to describe the appearance of the clothing in both the source and target videos. These person-specific and clothing-specific cues are combined with a randomly selected instruction template from the GPT-5.2-generated template pool and fed into Qwen-VL-8B to generate a precise and unambiguous editing instruction.

\textbf{Reference Construction}. Circle and bounding-box references are constructed following the same procedure as described in Sec.~\ref{object_addition_construction}. For clothing reference construction, we use Flux2-Klein-9B to synthesize the clothing reference image and employ GPT-5.2 to filter out failed cases.

\subsubsection{Human Swap}

For the human swap data, we selected 800K triplets of video, mask, and object name from the video pool.

\textbf{Editing Process}. We first used Qwen3-VL-8B to verify that the object name referred to a person. For each validated triplet, Qwen3-VL-8B was then used to generate a suitable replacement person whose appearance was distinct from that of the original person. Based on the generated person description, we constructed a text prompt and used a distilled human swap model to perform the human swap, producing edited videos at 720p resolution with 81 frames and 10 sampling steps. For the final paired data, the original video was treated as the target video, while the edited video was treated as the source video.

\textbf{Quality Filtering}. The quality filtering process is same as the object swap in Sec \ref{object_swap_construction}.

\textbf{Instruction Generation}. We use Qwen-VL-8B to generate unambiguous editing instructions that explicitly identify the edited object and avoid unclear references. Specifically, Qwen-VL-8B first determines which person is edited among all persons in the video. When multiple persons are present, it additionally describes the relative position of the edited person to enable precise localization. Next, Qwen-VL-8B extracts discriminative attributes of both the source and target persons, such as gender, age, appearance, clothing, and other visually salient characteristics, to further reduce referential ambiguity. Finally, these detected attributes are used by Qwen-VL-8B to compose a clear and specific editing instruction.

\textbf{Reference Construction}. Circle and bounding-box references are constructed following the same procedure as described in Sec.~\ref{object_addition_construction}. For human references, we use Flux2-Klein-9B to synthesize the corresponding human reference image and employ GPT-5.2 to filter out failed cases.

\subsubsection{Watermark Removal}\label{watermark_removal_construction}

For the watermark removal task, we constructed paired data by synthetically adding watermarks to clean videos.

\textbf{Editing Process}. We first used GPT-5.2 to generate 100 logo-generation prompts, which were then used by FLUX.2-Klein-9B~\cite{flux2_klein} to produce logos with white backgrounds. We removed the white backgrounds using OpenCV to obtain transparent logo background. The logo images can be regarded as the watermark images. Each watermark was inserted into a video at a randomly sampled position, with its scale sampled from $[0.2, 0.6]$ and opacity sampled from $[0.7, 1.0]$. The resulting watermarked video was used as the source video, while the original clean video was used as the target video. 

\textbf{Quality Filtering}. This part doesn't require quality filtering, since adding the watermark on the video couldn't result in bad quality such as artifacts and bad motion.

\textbf{Instruction Generation}. We use qwen-vl-8b to polish the instruction selected from the instruction pool generated by gpt5.2. 

\subsubsection{Subtitle Removal}

For the subtitle removal task, we constructed paired data by synthetically adding subtitles to clean videos. Specifically, we selected 300K triplets of video, mask, and caption from the video pool. 

\textbf{Editing Process}. We used OpenCV-based rendering algorithms to place the caption onto the video, with randomized font styles and layout parameters to increase diversity. The resulting subtitled video was used as the source video, while the original clean video was used as the target video. 

\textbf{Quality Filtering}. This part is same as the Sec \ref{watermark_removal_construction}.

\textbf{Instruction Generation}. This part is same as the Sec \ref{watermark_removal_construction}.

\subsection{Controllable Editing Video Pairs Construction Details}

For all sub-datasets, unless stated otherwise, editing instructions are constructed by instantiating predefined templates and then refined using Qwen-VL-8B.

\subsubsection{Colorization}

For the colorization task, we constructed paired data by converting clean videos into grayscale. Specifically, we randomly selected 300K pairs of video and caption from the video pool and applied grayscale conversion to the videos. All videos were processed at 720p resolution with 129 frames. The resulting grayscale video was used as the source video, while the original color video was used as the target video.

\subsubsection{Deblur}

For the deblur task, we constructed paired data by synthetically degrading clean videos with blur. Specifically, we randomly selected 300K pairs of video and caption from the video pool and applied Gaussian blur with varying parameters to generate blurred inputs. All videos were processed at 720p resolution with 129 frames. The resulting blurred video was used as the source video, while the original clean video was used as the target video.

\subsubsection{Upscale}

For the upscale task, we constructed paired data by synthetically degrading clean videos through downsampling. Specifically, we randomly selected 300K pairs of video and caption from the video pool, resized each video to a randomly sampled lower resolution, and then upsampled it back to 720p. All videos contained 129 frames. The resulting low-resolution-degraded video was used as the source video, while the original clean video was used as the target video.

\subsubsection{Outpainting}

For the outpainting task, we constructed paired data by synthetically cropping clean videos. Specifically, we randomly selected 300K pairs of video and caption from the video pool, cropped out the surrounding background region, and resized the remaining center region back to 720p. All videos contained 129 frames. The resulting cropped-and-resized video was used as the source video, while the original full video was used as the target video.

\subsubsection{Inpainting}

For the inpainting task, we constructed paired data by synthetically masking object regions in clean videos. Specifically, we randomly selected 300K triplets of video, masks, and object name from the video pool. For each video, we set the corresponding object region to black, with the mask randomly dilated by a few pixels to reduce overfitting to exact mask boundaries. All videos were processed at 720p resolution with 129 frames. The resulting masked video was used as the source video, while the original clean video was used as the target video.

\subsubsection{Hed to Video}

For the HED-to-video task, we constructed paired data by extracting edge maps from clean videos. Specifically, we randomly selected 300K pairs of video and caption from the video pool and applied an HED detector to generate the corresponding HED video. All videos were processed at 720p resolution with 129 frames. The resulting HED video was used as the source video, while the original clean video was used as the target video.

\subsubsection{Depth to Video}

For the depth-to-video task, we constructed paired data by extracting depth maps from clean videos. Specifically, we randomly selected 300K pairs of video and caption from the video pool and applied a MiDaS depth detector to generate depth videos at $512 \times 512$ resolution. The depth videos were then resized to 720p. All final videos contained 129 frames. The resulting depth video was used as the source video, while the original clean video was used as the target video.

\subsubsection{Canny to Video}

For the Canny-to-video task, we constructed paired data by extracting Canny edge maps from clean videos. Specifically, we randomly selected 300K pairs of video and caption from the video pool and applied a Canny detector to generate the corresponding Canny video. All videos were processed at 720p resolution with 129 frames. The resulting Canny video was used as the source video, while the original clean video was used as the target video.

\subsubsection{Fake Scribble to Video}

For the FakeScribble-to-video task, we constructed paired data by extracting simplified scribble-like structure maps from clean videos. Specifically, we randomly selected 300K pairs of video and caption from the video pool and applied a fake scribble detector to generate the corresponding fake scribble videos. The fake scribble detector was built upon the HED detector: it suppresses weak activations in the HED maps, binarizes the remaining strong responses, and further thickens the detected edges to produce scribble-like representations. All videos were processed at 720p resolution with 129 frames. The resulting fake scribble video was used as the source video, while the original clean video was used as the target video.

\subsubsection{Video to Hed/Depth/Canny/FakeScrible}

For the video-to-HED, video-to-depth, video-to-Canny, and video-to-FakeScribble tasks, we constructed paired data by reversing the source and target videos from the corresponding HED-to-video, depth-to-video, Canny-to-video, and FakeScribble-to-video datasets. Specifically, the original clean video was used as the source video, while the extracted condition video, i.e., the HED, depth, Canny, or FakeScribble video, was used as the target video. All videos were processed at 720p resolution with 129 frames.

\subsubsection{Video Object Detection}

For the video object detection task, we constructed paired data by converting instance annotations into detection videos. Specifically, we randomly selected 300K triplets of video, masks, and object names from the video pool. For each video, we assigned a different color to each object instance and set the background to black, thereby generating a video-level detection representation. All videos were processed at 720p resolution with 129 frames. The original video was used as the source video, while the resulting detection video was used as the target video.  Since this task only uses circle and bounding-box references, we follow the same construction procedure as described in Sec.~\ref{object_addition_construction}.

\section{Image Dataset Construction Details}\label{image_dataset_construction}

We construct multiple reference-based image editing subsets, including \textbf{style transfer}, \textbf{object addition}, \textbf{object retexture}, \textbf{object swap}, \textbf{human swap}, and \textbf{virtual try-on}. Each subset is organized as a quadruple consisting of a source image, a reference image, a target image, and an editing instruction. Across these subsets, the construction pipeline generally follows three stages: reference-conditioned target generation, source image derivation, and automatic instruction generation with visual quality filtering. Unless otherwise specified, GPT-5.2 is used to generate candidate concepts and instruction templates, Qwen3-VL-8B is used for prompt enhancement, instruction refinement, and visual verification, and Flux2-Klein-9B serves as the image generation and editing backbone. All images are created with 720P.

\subsection{Style Transfer}

We construct the image style transfer subset using a multi-stage generation and filtering pipeline.

\noindent\textbf{Generation Process.}
We first prompt GPT-5.2 to generate 200 diverse candidate style names. During triplet construction, we randomly sample one style name for each instance. Conditioned on the sampled style, Qwen3-VL-8B produces a detailed image-generation prompt, which is then fed into Flux2-Klein-9B to synthesize the stylized target image.

To obtain the corresponding source image, i.e., a realistic image without the target stylization, we apply a three-step inverse transformation. First, we convert the stylized target into a realistic image using the instruction ``\textit{make it realistic style}''. Second, we transform the resulting image into grayscale to suppress residual color and texture patterns. Third, we recolor the grayscale image with Flux2-Klein-9B using a new color specification generated by Qwen3-VL-8B. This procedure preserves the overall content and layout while reducing the original stylization cues.

In parallel, we construct the reference image by prompting Qwen3-VL-8B to extract representative texture and color attributes from the stylized target image. These attributes are combined with the sampled style name and a new content description to form a reference-generation prompt, which is then used by Flux2-Klein-9B to synthesize the reference image. The final training example therefore consists of a realistic source image, a style reference image, and a stylized target image.

\noindent\textbf{Instruction Generation.}
We use GPT-5.2 to build a diverse pool of style-transfer instruction templates. For each sample, the sampled style name is inserted into a randomly selected template to instantiate the initial instruction, which is further refined by Qwen3-VL-8B for clarity and naturalness.

\noindent\textbf{Quality Filtering.}
We employ Qwen3-VL-8B to filter invalid triplets. We discard samples where the reference image is overly similar to the target image in content, as such cases may leak target structure rather than only style information. We also remove samples where the source remains too similar to the target in color, texture, or style, since these examples provide weak supervision for reference-based style transfer.

\subsection{Object Addition}

We construct the reference-based object addition subset using a reference-to-target generation pipeline followed by object removal.

\noindent\textbf{Generation Process.}
We first prompt GPT-5.2 to generate 1,000 diverse object names. For each sample, an object name is randomly selected and used to synthesize a clean reference image, where the object is clearly presented on a white background. The generation prompt is refined by Qwen3-VL-8B and executed by Flux2-Klein-9B. Given this reference image, Flux2-Klein-9B composes a realistic target image that naturally contains the referenced object. We then apply Flux2-Klein-9B again to remove the inserted object from the target image, yielding the corresponding source image. Thus, each training sample contains a source image without the object, a reference image depicting the object to be added, and a target image where the object is present.

\noindent\textbf{Instruction Generation.}
Following the style-transfer subset, GPT-5.2 is used to generate a diverse set of object-addition instruction templates. For each sample, we randomly select one template and instantiate it with the object name. The resulting instruction is further polished by Qwen3-VL-8B to improve fluency, specificity, and visual grounding.

\noindent\textbf{Quality Filtering.}
We use Qwen3-VL-8B to conduct two visual consistency checks: whether the reference object appears in the target image, and whether the corresponding object has been removed from the source while preserving the remaining scene. Samples failing either check are discarded.

\subsection{Object Retexture}

We construct the object retexture subset following a similar reference-conditioned generation protocol, where the reference image specifies the desired texture rather than object identity.

\noindent\textbf{Generation Process.}
We prompt GPT-5.2 to generate 200 diverse texture categories, covering materials, patterns, and surface appearances. For each texture category, we synthesize a reference image in which the specified texture occupies the full image, ensuring that the reference provides a clean and unambiguous texture condition.

Given the reference texture image, Flux2-Klein-9B generates a target image containing an object whose surface follows the same texture as the reference. To construct the corresponding source image, we use Flux2-Klein-9B to recolor or weaken the textured object in the target image while preserving its identity, spatial layout, and background. This yields paired examples where the source contains the same object with the target texture removed or disrupted, while the target restores the reference-guided texture.

\noindent\textbf{Instruction Generation.}
We use GPT-5.2 to generate a pool of object-retexturing instruction templates. Each template is instantiated with the corresponding texture name and refined by Qwen3-VL-8B for naturalness, specificity, and alignment with the editing task.

\noindent\textbf{Quality Filtering.}
Qwen3-VL-8B verifies both texture consistency between the reference and target images and sufficient texture difference between the source and target images. Only samples where the target object matches the reference texture and the source object no longer exhibits the target texture are retained.

\subsection{Object Swap}

Object swap follows the same overall construction protocol as object addition, but differs in how the source image is derived. Instead of removing the referenced object from the target image, we replace it with another object.

\noindent\textbf{Generation Process.}
We first ask GPT-5.2 to generate 1,000 diverse object categories. For each sample, one category is randomly selected and used to synthesize a reference image with Flux2-Klein-9B, where the object is clearly presented as the visual condition. Given the reference image, Flux2-Klein-9B generates a realistic target image containing the same object in a natural scene.

To obtain the source image, we edit the target image by replacing the referenced object with a different object while preserving the scene layout, background, and other non-edited regions. The resulting sample therefore contains an alternative object in the source image, the desired object in the reference image, and the swapped-in object in the target image.

\noindent\textbf{Instruction Generation and Quality Filtering.}
Instruction generation follows the same template-based procedure as object addition, using object-swap templates generated by GPT-5.2 and refined by Qwen3-VL-8B. For quality filtering, Qwen3-VL-8B verifies that the target object matches the reference object and that the source contains a different object at the corresponding location while preserving the surrounding scene. Samples failing either criterion are removed.

\subsection{Human Swap}

Human swap instantiates the object-swap pipeline in the human domain, where the reference condition specifies a person rather than a generic object.

\noindent\textbf{Generation Process.}
We first prompt GPT-5.2 to enumerate diverse human categories, such as an old man, a young girl, and other demographic or appearance descriptions. For each sample, a human category is randomly selected and expanded by Qwen3-VL-8B into a detailed prompt. Flux2-Klein-9B then generates a reference image depicting the corresponding person.

Given the reference image, Flux2-Klein-9B synthesizes a realistic target image containing the same person in a natural scene. To derive the source image, we replace the target person with another plausible person while preserving the scene layout, background, pose context, and other non-edited regions. This forms a human identity replacement task, where the source contains an alternative person, the reference specifies the desired person, and the target contains the referenced person.

\noindent\textbf{Instruction Generation and Quality Filtering.}
We follow the same procedure as object swap, but use human-swap-specific instruction templates. The selected human category is inserted into the template and refined by Qwen3-VL-8B. During filtering, Qwen3-VL-8B checks whether the reference and target depict the same person and whether the source contains a different person while maintaining scene consistency. Only samples passing both checks are retained.

\subsection{Virtual Try-On}

The virtual try-on subset follows the same reference-conditioned editing protocol as human swap, but the reference condition specifies clothing rather than identity.

\noindent\textbf{Generation Process.}
We first prompt GPT-5.2 to enumerate 200 diverse clothing categories, including garments and accessories such as skirts, hats, coats, and other wearable items. For each sample, a clothing category is randomly selected and expanded into a detailed reference-generation prompt by Qwen3-VL-8B. Flux2-Klein-9B then synthesizes a clean reference image depicting the target clothing item.

Given the clothing reference, Flux2-Klein-9B generates a target image containing a person wearing the corresponding item. To obtain the source image, we replace the reference clothing in the target image with another appropriate garment suggested by Qwen3-VL-8B, while preserving the person, pose, background, and global scene layout. Each training example therefore consists of a source image where the person wears alternative clothing, a reference image specifying the desired clothing, and a target image where the person wears the reference clothing.

\noindent\textbf{Instruction Generation and Quality Filtering.}
Instruction generation follows the same template-based procedure as human swap, but with try-on-specific templates. The clothing category is inserted into the selected template and refined by Qwen3-VL-8B to better describe the clothing replacement task. For filtering, Qwen3-VL-8B verifies that the target person wears the reference clothing and that the source contains a different but plausible garment while preserving the person and scene. Samples failing either condition are discarded.

\sloppy
\section{Alignment Between MLLM-based Video Evaluation and Human Preference}\label{mllm_alignment}
\paragraph{Objective.}
We verify whether MLLM-based evaluators provide judgments consistent with human preference on instruction-guided and reference-guided video editing. The evaluation set contains 30 shared editing inputs and outputs from three systems for every input, yielding 90 videos with 30 from each system. The human annotator, GPT-5.5, and Gemini-3-Pro received the same evidence, including the editing instruction, source video, generated video, and the injected reference image when enabled by metadata. Screening information used to construct the set was hidden from all evaluators.

\paragraph{Human annotations.}
Each output was assigned a positive or negative label, a short rationale, and a confidence level. All 90 samples were completed. The latest annotations contain 35 positives and 55 negatives. Annotator confidence is high for 80 samples and medium for 10.

\begin{table}[t]
\centering
\caption{Automatic evaluation configuration. Source and output frames use matched relative timestamps.}
\label{tab:eval-config}
\small
\begin{tabular}{@{}>{\raggedright\arraybackslash}p{0.34\columnwidth}p{0.59\columnwidth}@{}}
\toprule
Setting & Value \\
\midrule
Evaluators & GPT-5.5 and Gemini-3.5-Pro \\
Video sampling & 8 frames at uniform interval midpoints \\
Spatial resolution & Longest image side 512 pixels \\
Visual input & Source frames, enabled reference image, output frames \\
Output & Binary label, rationale, confidence \\
Decision rule & Positive only if perfect or nearly perfect; partial completion is negative \\
Decoding & API-default temperature/top-$p$; no seed \\
Reliability & 300-s timeout and strict JSON validation \\
Run outcome & Both evaluators: 90/90 successful \\
\bottomrule
\end{tabular}
\end{table}

\paragraph{Evaluation criteria.}
The prompt asks the evaluator to jointly assess task completion, use of the reference identity/appearance/material, preservation of unedited content, temporal consistency, deformation and artifacts, and overall visual quality. The binary decision uses a strict criterion: only perfect or nearly perfect outputs are positive, while partial completion is negative.

\paragraph{Metrics.}
Human decisions are treated as the reference. We report binary-label accuracy, positive recall, negative specificity, positive precision, F1, and balanced accuracy. 

\begin{table}[t]
\centering
\caption{Binary-label agreement with human judgments over 90 samples. Bold indicates the better result between evaluators.}
\label{tab:overall}
\small
\begin{tabular}{@{}lrr@{}}
\toprule
Metric & GPT-5.5 & Gemini-3-Pro \\
\midrule
Accuracy & \best{94.44\%} & 92.22\% \\
Positive recall & 91.43\% & \best{97.14\%} \\
Negative specificity & \best{96.36\%} & 89.09\% \\
Positive precision & \best{94.12\%} & 85.00\% \\
F1 & \best{92.75\%} & 90.67\% \\
Balanced accuracy & \best{93.90\%} & 93.12\% \\
\bottomrule
\end{tabular}
\end{table}

\begin{table}[t]
\centering
\caption{Binary confusion matrices against human labels.}
\label{tab:confusion}
\small
\begin{tabular}{@{}lrrrrrr@{}}
\toprule
Evaluator & TP & FN & FP & TN & Pos. rec. & Neg. spec. \\
\midrule
GPT-5.5 & 32 & 3 & 2 & 53 & 91.43\% & 96.36\% \\
Gemini-3-Pro & 34 & 1 & 6 & 49 & 97.14\% & 89.09\% \\
\bottomrule
\end{tabular}
\end{table}

\paragraph{Results.}
GPT-5.5 agrees with the human binary preference on 85 of 90 videos (94.44\%), while Gemini agrees on 83 of 90 (92.22\%). GPT-5.5 has three false negatives and two false positives; Gemini has one false negative and six false positives. Thus Gemini is more sensitive to human-positive samples (97.14\% recall versus 91.43\%), whereas GPT-5.5 is more selective on human-negative samples (96.36\% specificity versus 89.09\%). The two MLLMs agree directly on 80 of 90 binary labels (88.89\%).

Overall, both MLLM evaluators are strongly aligned with human binary preference, with GPT-5.5 performing better on most aggregate classification measures and Gemini obtaining higher positive recall. Within this 90-sample study, label accuracies above 92\% support using MLLM-based evaluation for large-scale binary screening.

\section{User Study Details}
  \label{app:user-study}

  We recruited 33 participants to evaluate video editing results through a
  web-based questionnaire.

  The study comprised two sequential parts. Part~1 included 12 cases,
  each comparing the outputs of 15 methods. For each case, participants
  were shown a source video and a textual editing instruction. Part~2 also
  included 12 cases, each comparing three methods; participants were
  additionally provided with a reference image specifying the desired
  target appearance.

  For each case, participants answered forced-choice questions by selecting
  the best candidate from anonymized outputs labeled A, B, C, and so on.
  Part~1 evaluated four criteria, whereas Part~2 evaluated five criteria:

  \begin{itemize}
    \item \textbf{Text alignment} --- Which candidate best follows the
      textual editing instruction?
    \item \textbf{Visual quality} --- Which candidate exhibits the highest
      visual quality in terms of sharpness, artifact suppression, and realism?
    \item \textbf{Motion consistency} --- Which candidate demonstrates the
      most temporally consistent and natural motion?
    \item \textbf{Background preservation} --- Which candidate best preserves
      the unedited regions of the scene, including the background and
      unrelated objects?
    \item \textbf{Reference identity} --- Evaluated only in Part~2. Which
      candidate best preserves the identity and appearance of the subject
      specified by the reference image?
  \end{itemize}

\begin{figure*}[!htp]
    \centering
    \includegraphics[width=0.8\linewidth]{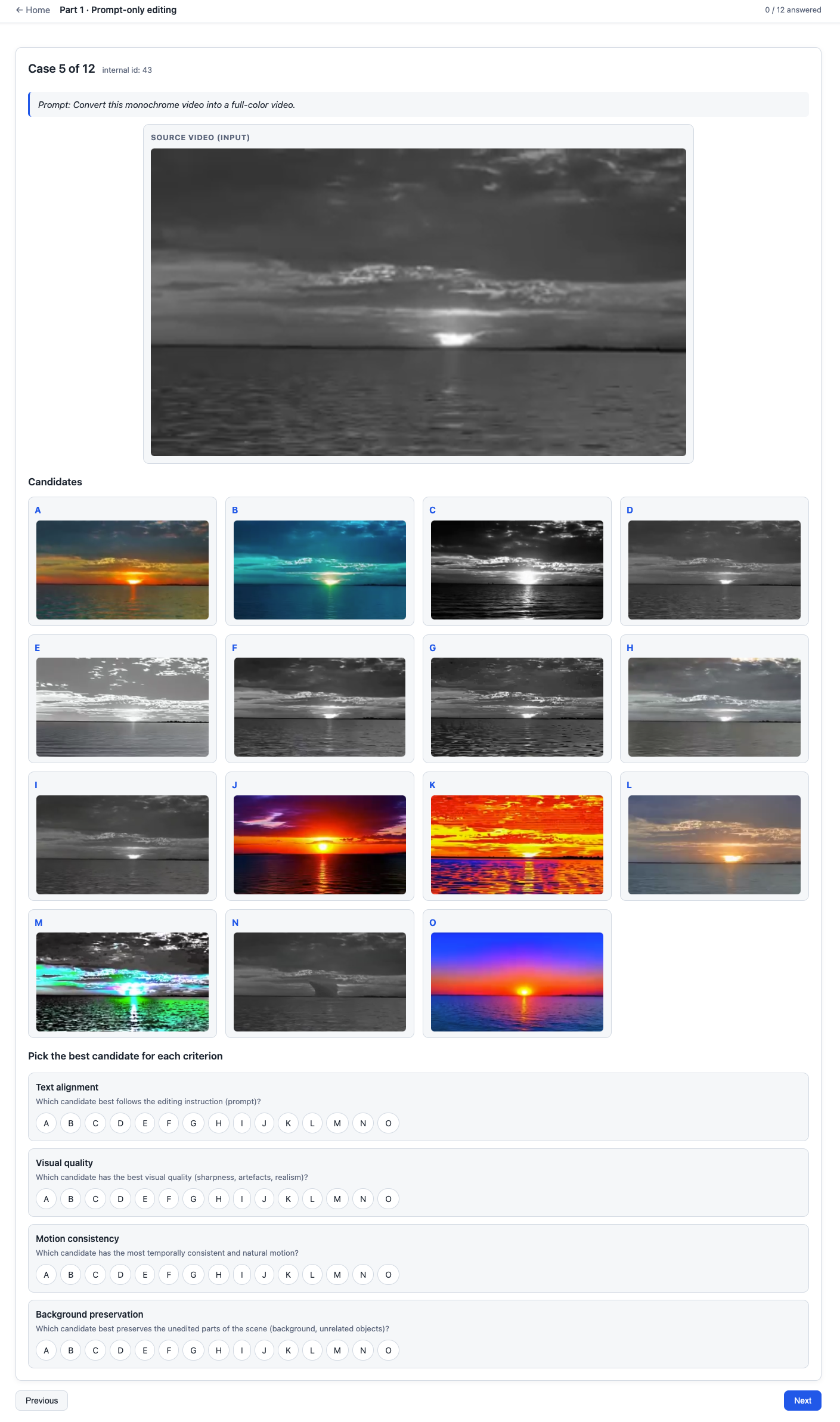}
    \caption{User-study interface for instruction-based video editing without visual references.}
    \label{fig:user_study_non_ref}
\end{figure*}

\begin{figure*}[!htp]
    \centering
    \includegraphics[width=0.925\linewidth]{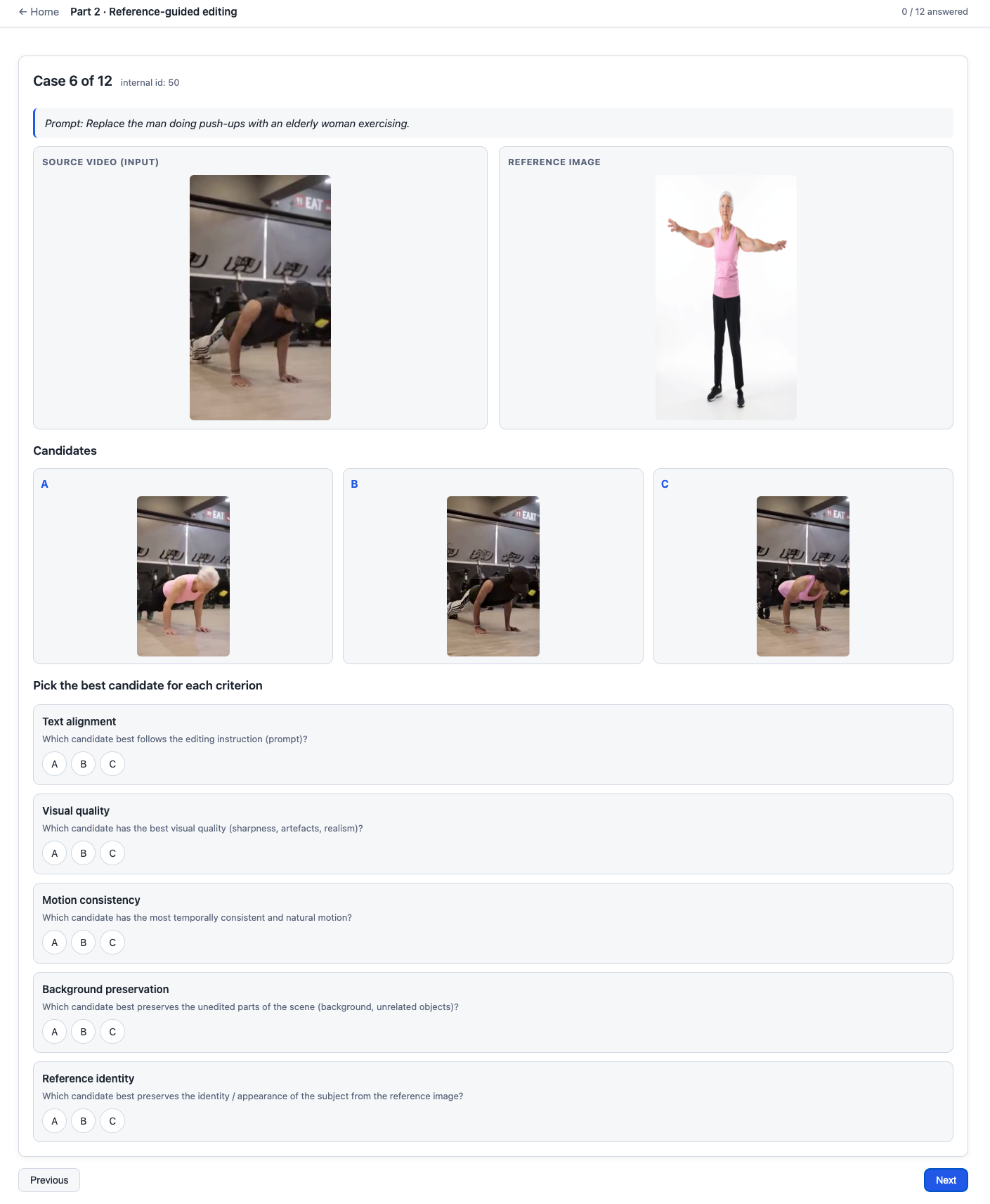}
    \caption{User-study interface for reference-based video editing with visual references.}
    \label{fig:user_study_ref}
\end{figure*}

The web pages for instruction‑based and reference‑guided video editing in the user study are in Figure \ref{fig:user_study_non_ref} and Figure \ref{fig:user_study_ref}.

\section{Details of the MLLM-Based Benchmark Evaluation}
\label{app:mllm_evaluation}

This section provides implementation details of the MLLM-as-judge protocol used in our benchmark. As described in the main paper, the benchmark consists of 100 instruction-based editing cases and 100 reference-based editing cases. The latter covers diverse forms of visual guidance, including object references, texture references, bounding boxes, and circle-marked regions. We use two independent MLLM judges, GPT-5.5 and Gemini-3-Pro, and apply the same input construction, question set, scoring rubric, and aggregation protocol to both judges.

\subsection{Evaluation Inputs}
\label{app:mllm_inputs}

For each edited video, we uniformly sample three frames in temporal order. Each frame is resized while preserving its aspect ratio before being passed to the judge. The MLLM receives (1) the editing instruction, (2) the sampled frames from the edited video, and, when applicable, (3) the corresponding visual reference. Depending on the editing task, the visual reference can be an object image, a texture image, a bounding-box annotation, or a circle-marked image. If more than one reference file is available, the implementation selects the object or texture reference first, followed by the bounding-box reference and then the circle-marked reference.

The evaluation questions comprise two parts. First, task-specific questions are constructed according to the editing instruction and task type. They assess whether the requested operation has been performed, whether the target attributes match the instruction or reference, whether unrelated regions are preserved, and whether the overall result follows the instruction. Second, two task-agnostic questions are appended to every case to assess visual artifacts and unintended objects. This combination allows the protocol to measure both task fulfillment and general output quality.

\subsection{Prompt Presented to the MLLM Evaluation}
\label{app:mllm_prompt}

The complete prompt template is shown below. Text enclosed in angle brackets is instantiated separately for each test case. The reference-image block is omitted for instruction-based cases that do not require visual guidance, and each image placeholder is replaced by the corresponding image input in the actual multimodal request.

\begin{tcolorbox}[
    colback=white,
    colframe=black!75!white,
    title={MLLM-as-Judge Prompt},
    breakable
]
\small
You will evaluate an edited video. The edit instruction was: ``\emph{$<$EDITING INSTRUCTION$>$}.''

\medskip
\emph{[Included only when a visual reference is available:]}

Below is the reference image (showing the desired target / mask / circle):

$<$REFERENCE IMAGE$>$

\medskip
Below are $<$N$>$ sampled frames from the edited video, in order:

$<$EDITED VIDEO FRAME 1$>$

$<$EDITED VIDEO FRAME 2$>$

$<$EDITED VIDEO FRAME 3$>$

\medskip
Please answer the following questions about the edited video. For each question, give an integer score from 1 to 5 (1 = worst / does not satisfy at all, 5 = best / perfectly satisfies) and a brief comment. Return the result STRICTLY as a JSON array with one object per question, in order, with fields `score' (int 1--5) and `comments' (string). Do not include any text outside the JSON array.

\medskip
Questions:

1. $<$TASK-SPECIFIC QUESTION 1$>$

2. $<$TASK-SPECIFIC QUESTION 2$>$

$\cdots$

$K$. Are there visible artifacts in the edited video (e.g., distortions, flickering, broken textures, ghosting, harsh edges)?

$K+1$. Are there any objects that should NOT appear in the edited video (e.g., spurious or unintended objects introduced by the edit)?

\medskip
Output example:

[{``score'': 4, ``comments'': ``...''}, {``score'': 3, ``comments'': ``...''}]
\end{tcolorbox}

The task-specific questions are adapted to the requested edit rather than using only a single generic instruction-following question. For example, the following question set is used for a reference-guided texture-editing case whose instruction is ``Retexture the coast into black basalt.''

\begin{tcolorbox}[
    colback=white,
    colframe=black!75!white,
    title={Example Task-Specific Questions: Reference-Based Editing},
    breakable
]
\small
1. Rate whether the target (coast) is retextured as requested.

2. Rate whether the material or surface appearance clearly matches: black basalt.

3. Rate whether the edited result matches the provided reference image for the requested texture.

4. Rate whether the background and unrelated non-target areas remain unchanged, except for the requested edit.

5. Rate the overall instruction-following quality of this edited video.
\end{tcolorbox}

For instruction-based editing, the questions are similarly customized to the target object, requested operation, and desired attributes. For example, for the instruction ``Replace the chick with a ginger kitten,'' the task-specific questions are:

\begin{tcolorbox}[
    colback=white,
    colframe=black!75!white,
    title={Example Task-Specific Questions: Instruction-Based Editing},
    breakable
]
\small
1. Rate whether the chick has been replaced by a cat.

2. Rate whether the cat looks like a ginger kitten.

3. Rate whether all non-edited areas outside the chick/ginger kitten preserve their original color and structure.
\end{tcolorbox}

These task-specific questions are followed by the same two task-agnostic questions shown in the complete prompt template. Both MLLMs are required to return a JSON array containing one score and one brief justification for each question. Responses that cannot be parsed as a JSON array, contain an incorrect number of entries, or include invalid scores are retried; unsuccessful cases are marked as invalid rather than silently included in the aggregate.

\subsection{Scoring and Aggregation}
\label{app:mllm_aggregation}

Each question is scored on a five-point scale, where 1 indicates the poorest fulfillment and 5 indicates the best fulfillment. The questions are mapped to five evaluation dimensions:
\begin{itemize}
    \item \textbf{Instruction Alignment}: whether the requested editing operation is completed;
    \item \textbf{Attribute Alignment}: whether the edited target matches the requested attributes or visual reference;
    \item \textbf{Background Preservation}: whether non-target content remains unchanged;
    \item \textbf{Visual Quality}: whether the result is free of visible artifacts; and
    \item \textbf{No Undesired Object}: whether the edit avoids introducing unintended content.
\end{itemize}

For each method and dimension, we average the valid scores of all questions mapped to that dimension. We report the results from GPT-5.5 and Gemini-3-Pro separately, thereby avoiding an implicit preference for either judge. Per-video predictions, question-level scores, and textual justifications are retained to support reproducibility and qualitative error analysis.

\section{Construction of Stylization Prompts}
We use the same procedure to construct prompts for both first-frame stylization and video stylization. First, select a target style description $s$ from the style template library. Then, sample an instruction phrase $t$ from a set of semantically equivalent style‑transfer templates, which include \texttt{make it}, \texttt{make it into}, \texttt{transfer it to}, \texttt{transform it into}, \texttt{convert it to}, and \texttt{stylize it as}. Place the instruction $t$ before the selected style description $s$. The following list contains all 241 source style descriptions in file order.

\begin{tcolorbox}[
    colback=white,
    colframe=black!75!white,
    title={Example Task-Specific Questions: Instruction-Based Editing},
    breakable
]
\small
traditional Chinese ink painting style, expressive ink wash, classical East Asian\\
ukiyo-e style, Japanese woodblock, flat graphic\\
illuminated manuscript style, medieval decoration, intricate borders\\
voynich manuscript style, mysterious diagrams, archaic manuscript\\
renaissance style, classical European, humanistic realism\\
baroque style, dramatic lighting, ornate composition\\
rococo style, elegant ornament, playful decoration\\
romanticism style, emotional drama, historical scenes\\
impressionism style, soft brushstrokes, luminous color\\
post-impressionism style, bold color, structural brushwork\\
fauvism style, vivid colors, expressive shapes\\
cubism style, geometric fragmentation, abstract forms\\
abstract art style, nonrepresentational, expressive shapes\\
abstract expressionism style, gestural painting, emotional intensity\\
surrealism style, dreamlike, symbolic imagery\\
op art style, optical illusion, vibrating patterns\\
constructivist style, geometric abstraction, industrial modernism\\
bauhaus style, functional minimal, modernist design\\
art nouveau style, flowing lines, decorative floral MoTifs\\
neo-expressionism style, raw brushwork, emotional distortion\\
dada / conceptual style, experimental, idea-driven\\
folk art style, naive drawing, rustic charm\\
ethnic folk art style, traditional patterns, cultural MoTifs\\
quilted textile art style, stitched patterns, layered fabric\\
graffiti art style, vibrant spray paint, urban street\\
graphic design style, bold layout, clear typography\\
japanese graphic design poster style, minimal composition, sharp contrast\\
risograph print style, limited palette, grainy texture\\
stock illustration style, clean vector, commercial friendly\\
fashion illustration style, elongated figures, runway aesthetic\\
book page illustration style, text integrated, decorative layout\\
collage / montage style, cut-out layers, mixed imagery\\
glitch art style, pixel distortion, digital noise\\
vaporwave style, pastel neon, retro digital nostalgia\\
retro futurism style, 1950s sci-fi, bold colors\\
fantasy map style, hand-drawn cartography, parchment texture\\
manga style, stylized lines, dynamic panels\\
anime style, vibrant cel shading, expressive characters\\
chibi anime style, super deformed, cute proportions\\
children’s illustration style, gentle colors, storybook charm\\
pixel art style, 8-bit blocks, retro game look\\
90s video game style, chunky pixels, arcade nostalgia\\
low poly 3d style, simplified geometry, pastel minimalism\\
film photography style, grainy texture, analog mood\\
documentary photography style, natural world, candid realism\\
nature photography style, high detail, natural light\\
realism style, accurate proportions, detailed depiction\\
hyperrealism style, ultra detailed, lifelike rendering\\
photorealism style, camera-like, precise lighting\\
cinematic style, dramatic composition, atmospheric lighting
black and white style, grayscale tones, strong contrast\\
noir style, deep shadows, vintage detective mood\\
3d cgi style, volumetric forms, digital rendering\\
cgsociety style, polished cgi, high detail\\
unity creations style, stylized cgi, polished rendering\\
subsurface scattering style, translucent skin, realistic light\\
ink illustration style, bold linework, expressive strokes\\
color ink on paper style, fluid pigment, vivid color\\
watercolor painting style, soft washes, bleeding edges\\
doodle sketch style, spontaneous lines, playful simplicity\\
architectural sketching style, perspective lines, structural detail\\
interior design style, curated decor, spatial harmony\\
brutalist style, raw concrete, heavy geometry\\
victorian style, ornate decoration, historical detailing\\
concept art style, exploratory designs, cinematic mood\\
character concept art style, hero-focused, detailed costumes\\
weapon design style, functional fantasy, stylized shapes\\
game scene graph style, layered elements, structured game world\\
anatomical study style, partial anatomy, structural focus\\
gothic dark style, moody medieval, dramatic atmosphere\\
dark fantasy style, eerie creatures, occult ambience\\
epic fantasy style, grand landscapes, mythical beings\\
steampunk style, retro machinery, victorian futurism\\
cyberpunk style, neon city, dystopian tech\\
futuristic sci-fi style, advanced tech, sleek designs\\
vintage style, aged texture, retro palette\\
retro dark vintage gothic style, aged horror, moody retro\\
pulp noir style, high contrast, hard-boiled crime\\
stained glass window style, colored glass, luminous patterns\\
surrealist symbolic style, dreamlike scenes, strange juxtapositions\\
minimalist art style, clean forms, negative space\\
minimalist design style, clean lines, muted colors\\
isometric perspective style, 3d illusion, axonometric view\\
pop art style, bold colors, comic aesthetics\\
superflat style, flat planes, pop imagery\\
breath of the wild style, painterly fantasy, open world\\
warframe style, sci-fi ninja, biomechanical armor\\
pokémon style, cute creatures, anime game look\\
league of legends style, fantasy moba, stylized champions\\
afk arena style, fantasy heroes, mobile rpg\\
cookierun kingdom style, cute cookies, colorful fantasy\\
apex legends style, fast-paced fps, hero shooter\\
the elder scrolls style, high fantasy, open world rpg\\
fromsoftware style, dark fantasy, punishing atmosphere\\
detroit become human style, near-future, realistic sci-fi\\
studio ghibli style, warm fantasy, cozy worlds\\
makoto shinkai style, vivid skies, cinematic drama\\
kyoto animation (kyoani) style, soft colors, expressive characters\\
jojo’s bizarre adventure style, bold poses, dramatic lines\\
a silent voice (koe no katachi) style, emotional drama, grounded anime
soejima shigenori style, stylish characters, persona-like design\\
yoji shinkawa style, brushy ink, mecha silhouettes\\
akira toriyama style, cartoony lines, shonen energy\\
akihiko yoshida style, ornate costumes, jrpg fantasy\\
hong soonsang style, appealing shapes, animation-ready characters\\
michelangelo style, heroic anatomy, renaissance sculpture\\
claude monet style, soft light, impressionist landscapes\\
paul cézanne style, structured brushwork, proto-cubist forms\\
van gogh style, swirling brushstrokes, post-impressionism\\
mark rothko style, color fields, meditative abstraction\\
paul klee style, playful symbols, bauhaus abstraction\\
picasso style, cubist fragmentation, bold experimentation\\
mondrian style, primary colors, geometric grids\\
renoir style, soft figures, intimate impressionism\\
rembrandt style, chiaroscuro, emotional portraits\\
magritte style, deadpan surrealism, visual paradoxes\\
lichtenstein style, ben-day dots, comic pop art\\
dalí style, melting forms, hyperreal surrealism\\
botticelli style, graceful figures, mythological scenes\\
murakami takashi style, flat pop, superflat icons\\
kandinsky style, abstract geometry, spiritual color\\
collishaw style, immersive installations, atmospheric digital\\
kusama yayoi style, polka dots, infinite repetition\\
igor morski style, symbolic collages, surreal fantasy\\
jerry pinkney style, rich watercolor, children’s books\\
beatrix potter style, animal characters, classic storybook\\
jon klassen style, simple shapes, dry humor\\
kay sage style, architectural surrealism, silent landscapes\\
jeffrey catherine jones style, painterly fantasy, moody figures\\
yaacov agam style, kinetic optical, shifting patterns\\
david hockney style, bright color, contemporary pop\\
victor moscoso style, psychedelic posters, vibrating color\\
stefan koid style, contemporary fine art, experimental imagery\\
sui ishida style, dark manga, emotional characters\\
john harris style, atmospheric sci-fi, distant vistas\\
junji ito style, body horror, detailed linework\\
anton pieck style, fairy-tale towns, nostalgic detail\\
carl barks style, classic comics, duck characters\\
alphonse mucha style, art nouveau, ornate posters\\
andy warhol style, celebrity silkscreen, pop repetition\\
banksy style, political stencil, street satire\\
francisco de goya style, dark romanticism, historical scenes\\
diego rivera style, monumental murals, social themes\\
marc chagall style, dreamlike figures, floating compositions\\
edgar degas style, dancers, interior scenes\\
eugène delacroix style, dynamic brushwork, dramatic scenes\\
francis bacon style, distorted figures, raw emotion\\
frida kahlo style, symbolic self-portraits, surreal realism\\
gerald brom style, dark fantasy, gothic characters\\
gustav klimt style, gold leaf, decorative patterns
henri matisse style, bold color, simplified forms\\
j.m.w. turner style, luminous seascapes, atmospheric haze\\
jack kirby style, dynamic poses, cosmic comics\\
jackson pollock style, drip painting, gestural abstraction\\
johannes vermeer style, quiet interiors, soft light\\
jean-michel basquiat style, graffiti marks, neo-expressionism\\
marcel duchamp style, conceptual readymades, avant-garde\\
adrian donohue style, staged scenes, surreal photography\\
adrian tomine style, clean lines, contemporary comics\\
cai guo-qiang style, gunpowder drawings, explosive installations\\
drew struzan style, detailed portraits, movie posters\\
hans arp style, biomorphic shapes, dada abstraction\\
ilya kuvshinov style, big-eyed portraits, manga-inspired\\
james jean style, intricate lines, dreamlike illustration\\
jasmine becket-griffith style, big-eyed girls, pop surrealism\\
jean giraud MoTbius style, airy lines, visionary sci-fi\\
dennis stock style, classic b\&w, iconic portraits\\
michal lisowski style, moody lighting, cinematic digital art\\
paul lehr style, retro sci-fi, cosmic vistas\\
ross tran style, vibrant colors, energetic characters\\
swoon style, cut-paper portraits, street installations\\
tasha tudor style, nostalgic countryside, delicate linework\\
tintoretto style, dramatic poses, late renaissance\\
theodore robinson style, american impressionism, dappled light\\
titian style, rich color, renaissance figures\\
wlop style, luminous atmospheres, digital fantasy\\
yanjun cheng style, painterly portraits, soft realism\\
alena aenami style, glowing horizons, atmospheric mood\\
anton fadeev style, dynamic designs, concept art\\
charlie bowater style, character-focused, painterly digital\\
cory loftis style, expressive shapes, animation concept\\
fenghua zhong style, brushy strokes, heroic fantasy\\
greg rutkowski style, detailed fantasy, dynamic lighting\\
hong soonsang style, appealing characters, animation concept\\
osamu tezuka style, classic manga, round forms\\
rob gonsalves style, illusionary architecture, magic realism\\
sol lewitt style, conceptual structures, minimal geometry\\
yusuke murata style, high-energy action, polished manga\\
antonio mora style, double exposure, surreal portraits\\
yoji shinkawa style, gestural ink, military mecha\\
soejima shigenori style, fashion-forward, persona-like characters\\
yamada akihiro style, intricate lines, dark fantasy\\
munashichi style, dreamy cities, atmospheric scenery\\
suprematism style, flat geometric forms, spiritual abstraction\\
orphism style, circular color rhythms, lyrical abstraction\\
synchromism style, color harmonies, musical abstraction\\
rayonism style, radiant rays, fractured light\\
precisionism style, industrial architecture, crisp geometry\\
metaphysical painting style, long shadows, silent plazas\\
art deco style, luxurious ornament, streamlined geometry\\
vienna secession style, ornate lines, symbolist MoTifs
nabis school style, flat decorative color, intimate scenes\\
der blaue reiter style, spiritual color, expressive forms\\
new objectivity style, sharp realism, social commentary\\
arte povera style, humble materials, conceptual gestures\\
fluxus style, performance based, playful anti-art\\
gutai style, action painting, raw material experiments\\
situationist style, urban dérive, subversive collage\\
cobra group style, spontaneous brushwork, childlike figures\\
lowbrow art style, pop surrealism, underground comics\\
psychedelic art style, intense colors, swirling patterns\\
afrofuturism style, black cultural futurism, cosmic technology\\
solarpunk style, lush greenery, optimistic eco-future\\
dieselpunk style, smoky engines, interwar retro tech\\
atompunk style, mid-century rockets, atomic age optimism\\
memphis design style, clashing colors, playful geometry\\
international typographic style, clean grids, sans serif\\
y2k aesthetic style, chrome interfaces, tech nostalgia\\
frutiger aero aesthetic style, glossy gradients, sky imagery\\
weirdcore style, uncanny nostalgia, distorted reality\\
dreamcore style, soft surreal scenes, hazy atmosphere\\
cottagecore style, rural romance, soft pastels\\
goblincore style, messy nature, earthy clutter\\
liminal space photography style, empty corridors, uneasy calm\\
lomography style, heavy vignette, saturated colors\\
cyanotype photography style, deep blue tones, contact prints\\
wet plate collodion style, antique blur, silver tones\\
encaustic painting style, wax layers, textured surfaces\\
impasto painting style, thick paint, sculpted strokes\\
sgraffito mural style, scratched layers, contrasting colors\\
woodcut printmaking style, bold carving, high contrast\\
linocut printmaking style, graphic shapes, strong outlines\\
stippling ink illustration style, tiny dots, gradual shading, text characters, monospaced grid\\
generative algorithmic art style, coded visuals, procedural patterns\\
data moshing video art style, broken compression, smeared motion\\
demoscene graphics style, low-level code, retro effects\\
rubber hose animation style, stretchy limbs, vintage cartoon\\
ligne claire comic style, clean outlines, flat colors\\
papercut silhouette art style, crisp edges, layered paper\\
shadowbox diorama style, layered depth, theatrical lighting\\
persian miniature painting style, jewel tones, ornate patterns
\end{tcolorbox}

\begin{figure*}[htp]
    \centering
    \includegraphics[width=0.92\textwidth]{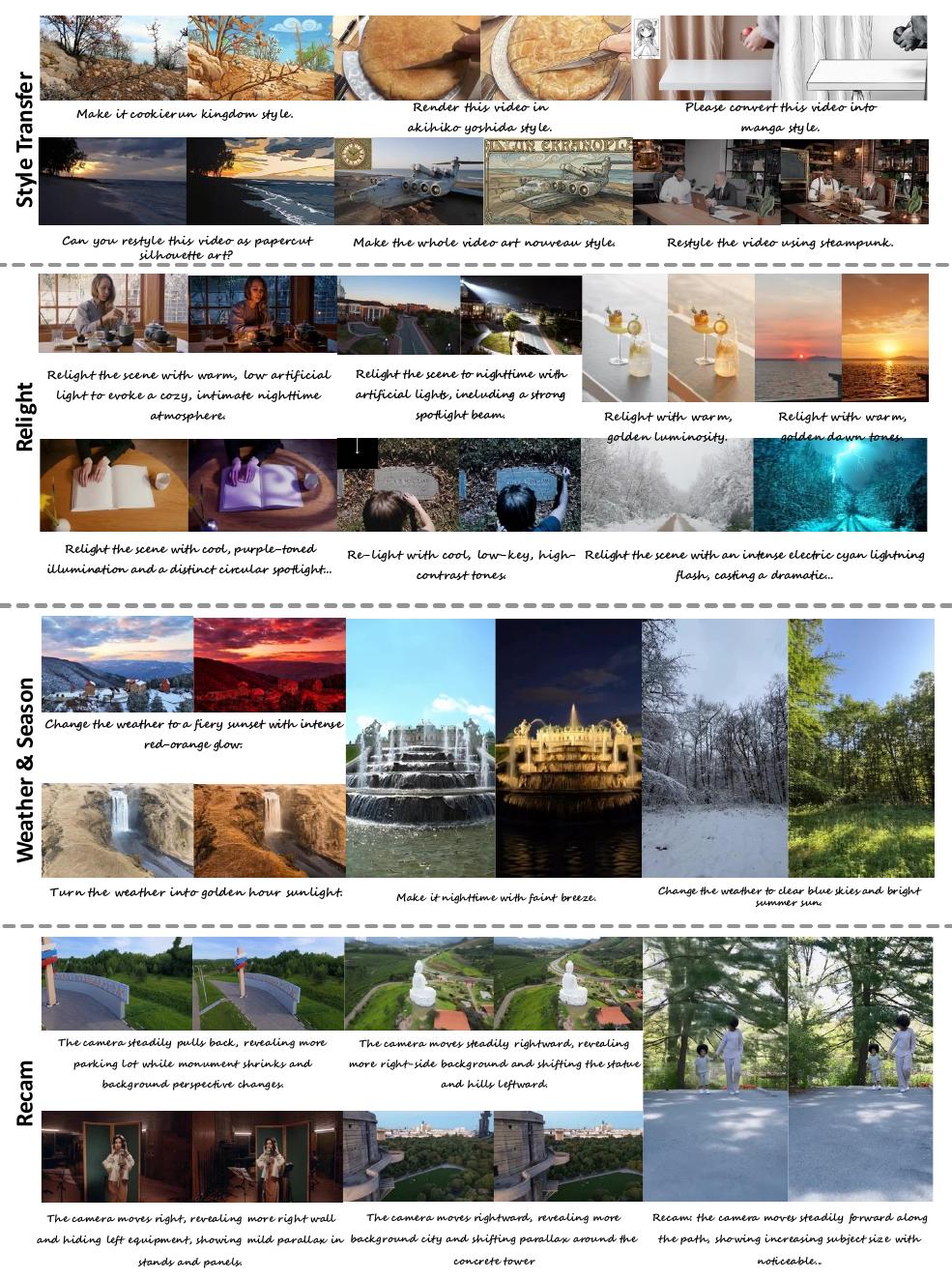}
    \caption{Visualization of style transfer, relighting, weather/season editing, and camera motion instructions. This figure summarizes prompts that modify the global appearance, illumination, environmental condition, or viewpoint trajectory of a video.}
    \label{fig:appendix-style-relight-weather-recam}
\end{figure*}

\begin{figure*}[htp]
    \centering
    \includegraphics[width=0.92\textwidth]{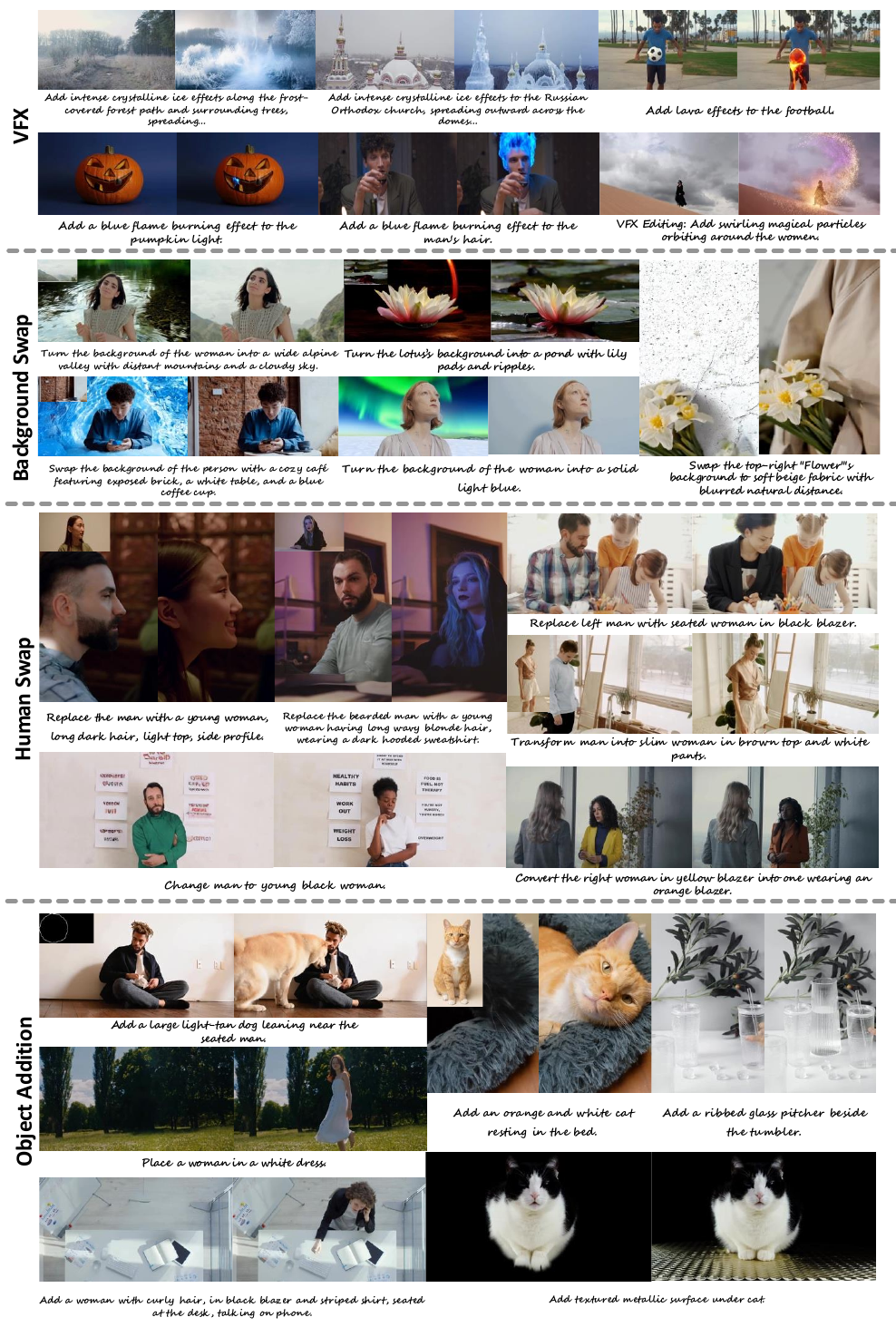}
    \caption{Visualization of VFX editing, background swap, human swap, and object addition instructions. These examples demonstrate how the dataset covers local visual effects, scene replacement, identity or person transformation, and insertion of new objects.}
    \label{fig:appendix-vfx-swap-addition}
\end{figure*}

\begin{figure*}[htp]
    \centering
    \includegraphics[width=0.92\textwidth]{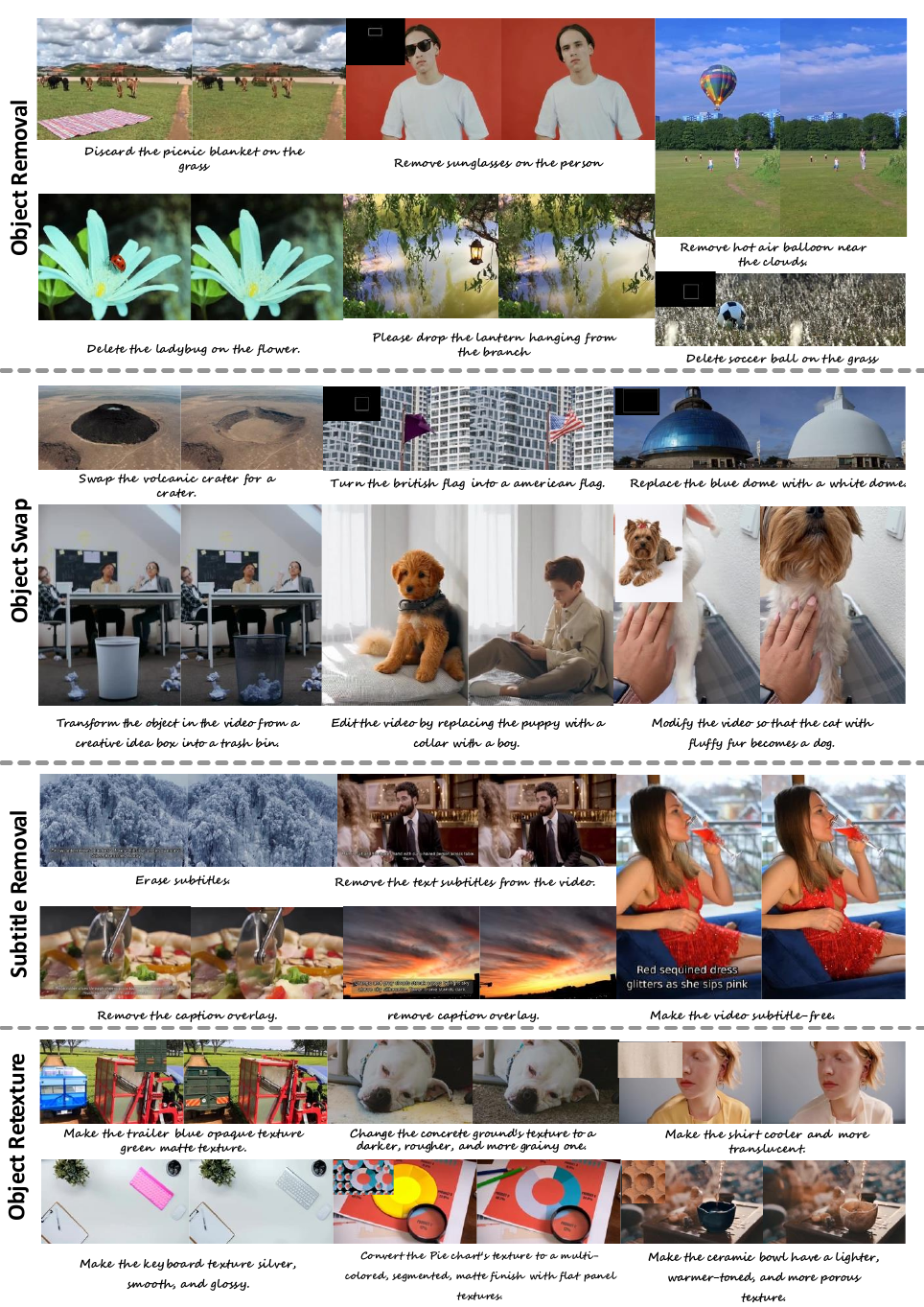}
    \caption{Visualization of object removal, object swap, subtitle removal, and object retexture instructions. The figure illustrates editing tasks that remove distracting elements, replace target objects, erase text overlays, or alter material appearance.}
    \label{fig:appendix-removal-swap-retexture}
\end{figure*}

\begin{figure*}[htp]
    \centering
    \includegraphics[width=0.92\textwidth]{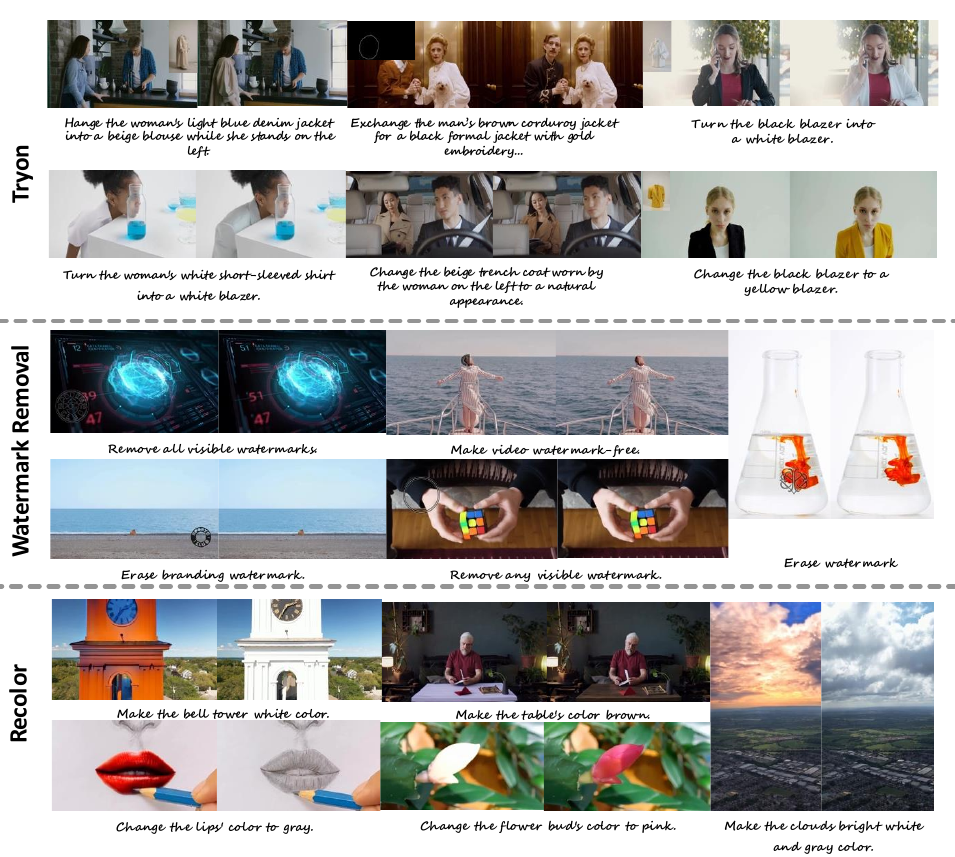}
    \caption{Visualization of virtual try-on, watermark removal, and recoloring instructions. These categories focus on changing clothing attributes, removing visible watermarks or branding, and modifying the color of selected regions or objects.}
    \label{fig:appendix-tryon-watermark-recolor}
\end{figure*}

\begin{figure*}[htp]
    \centering
    \includegraphics[width=0.92\textwidth]{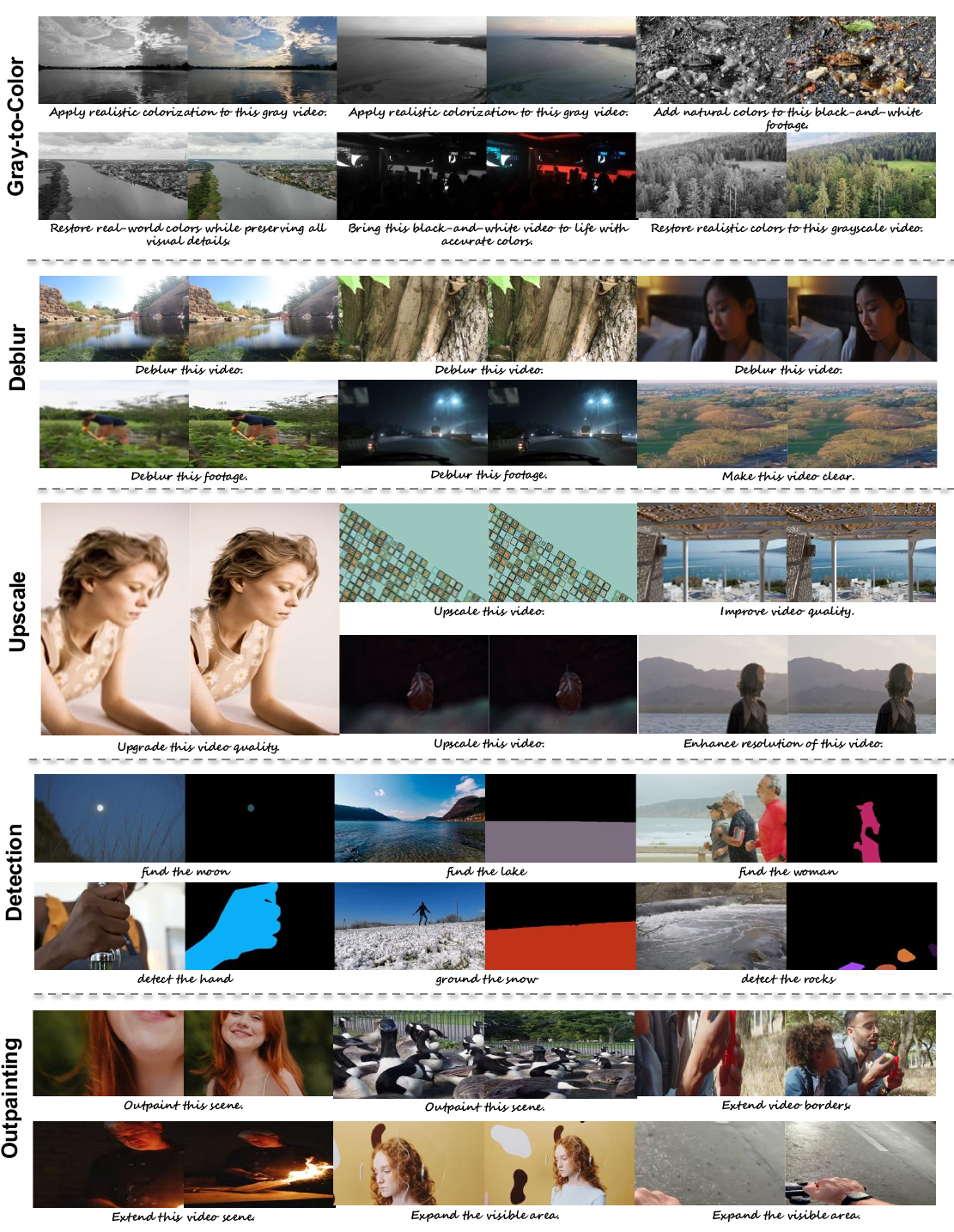}
    \caption{Visualization of video restoration, detection, and outpainting instructions. The examples include upscaling, colorizing grayscale videos, deblurring, grounding or detecting target entities, and expanding the visible scene beyond the original frame.}
    \label{fig:appendix-restoration-detection-outpainting}
\end{figure*}

\begin{figure*}[htp]
    \centering
    \includegraphics[width=0.92\textwidth]{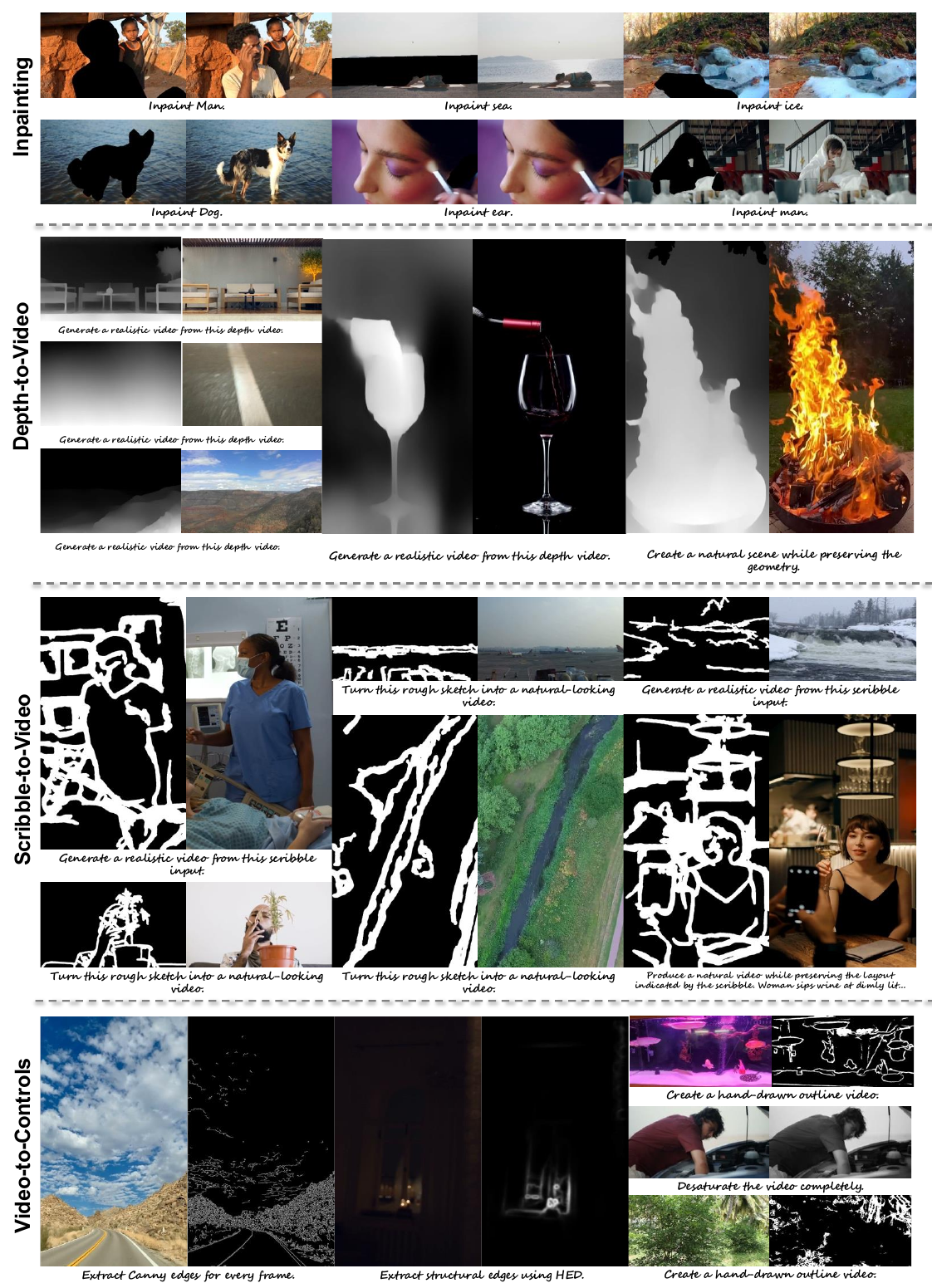}
    \caption{Visualization of inpainting and structure-conditioned video generation instructions. This figure covers masked region completion, depth-to-video generation, scribble-to-video generation, and conversion from videos to structural controls.}
    \label{fig:appendix-inpainting-structural-controls}
\end{figure*}

\begin{figure*}[htp]
    \centering
    \includegraphics[width=0.92\textwidth]{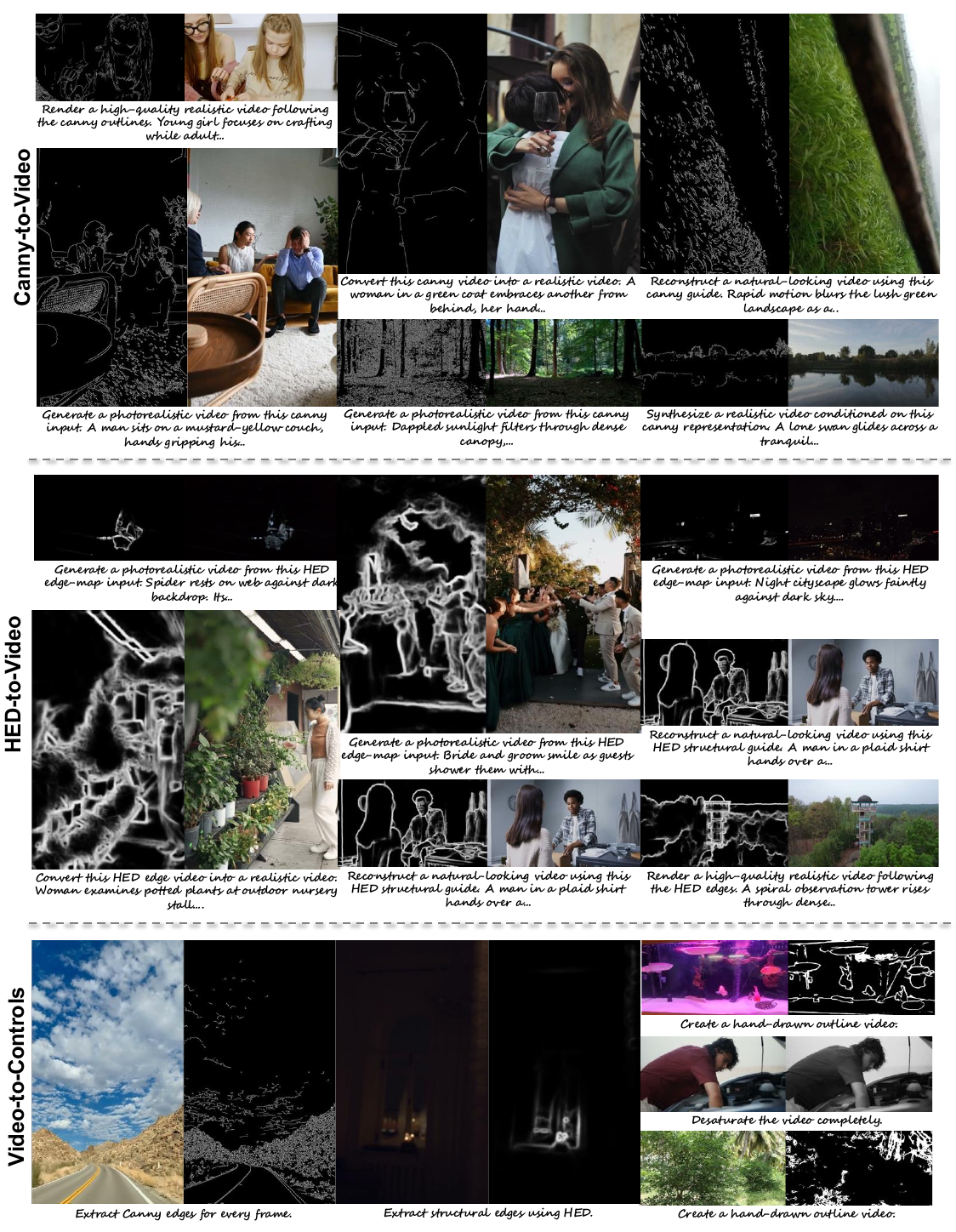}
    \caption{Visualization of Canny-to-video, HED-to-video, and video-to-control instructions. These examples highlight control-based generation, where edge maps or structural guides are used to synthesize realistic videos or extract control representations from videos.}
    \label{fig:appendix-canny-hed-controls}
\end{figure*}

\end{document}